\documentclass[12pt]{article}

\usepackage[utf8]{inputenc}

\usepackage{amsmath}
\usepackage{times}
\usepackage{graphicx}
\usepackage{color}
\usepackage{bm}

\usepackage{multirow}
\usepackage{csquotes}
\usepackage[style=apa,backend=biber,language=american]{biblatex}
\DeclareLanguageMapping{american}{american-apa}
 
\usepackage{multirow}
\usepackage{adjustbox}
\usepackage{booktabs}

\usepackage{rotating}
\usepackage{bbm}
\usepackage{latexsym}

\usepackage{csquotes}

\usepackage[modulo]{lineno}

\newcommand{\captionfonts}{\normalsize}

\usepackage{graphicx,kantlipsum,setspace}
\usepackage{caption}
\usepackage{hyperref}

\makeatletter  
\long\def\@makecaption#1#2{%
  \vskip\abovecaptionskip
  \sbox\@tempboxa{{\captionfonts #1: #2}}%
  \ifdim \wd\@tempboxa >\hsize
    {\captionfonts #1: #2\par}
  \else
    \hbox to\hsize{\hfil\box\@tempboxa\hfil}%
  \fi
  \vskip\belowcaptionskip}
\makeatother   

\usepackage{algorithm}
\usepackage{algpseudocode}
\newcommand{\pluseq}{\mathrel{+}=}

\DeclareFontFamily{U}{mathx}{\hyphenchar\font45}
\DeclareFontShape{U}{mathx}{m}{n}{<-> mathx10}{}
\DeclareSymbolFont{mathx}{U}{mathx}{m}{n}
\DeclareMathAccent{\widebar}{0}{mathx}{"73}

\newcommand{\alignedequation}[2]{ %
    \begin{equation} \label{#1} %
    \mbox{\fontsize{9}{9}\selectfont\(\begin{aligned}#2\end{aligned}\)} %
    \end{equation} %
} 

\begin{document}
\hspace{13.9cm}

\ \vspace{20mm}\\
%
{\LARGE A Novel Predictive-Coding-Inspired Variational RNN Model for Online Prediction and Recognition}
\ \\
\ \\
{\bf \large Ahmadreza Ahmadi$^{\displaystyle 1, \displaystyle 2}$~~Jun Tani$\footnote[1]{Corresponding author}^{\displaystyle 1}$}\\
{$^{\displaystyle 1}$Okinawa Institute of Science and Technology, Okinawa, Japan 904-0495.}\\
{$^{\displaystyle 2}$School of Electrical Engineering, Korea Advanced Institute of Science
and Technology, Daejeon, 305-701, Republic of Korea.}\\
%
{\bf Keywords:} Recurrent neural network, variational Bayes, predictive coding, generative model, inference model
\thispagestyle{empty}
\markboth{}{NC instructions}
\ \vspace{-0mm}\\
%
\begin{center} {\bf Abstract} \end{center}

This study introduces PV-RNN, a novel variational RNN inspired by the predictive-coding ideas. The model learns to extract the probabilistic structures hidden in fluctuating temporal patterns by dynamically changing the stochasticity of its latent states. Its architecture attempts to address two major concerns of variational Bayes RNNs: how can latent variables learn meaningful representations and how can the inference model transfer future observations to the latent variables. PV-RNN does both by introducing adaptive vectors mirroring the training data, whose values can then be adapted differently during evaluation. Moreover, prediction errors during backpropagation---rather than external inputs during the forward computation---are used to convey information to the network about the external data. For testing, we introduce \emph{error regression} for predicting unseen sequences as inspired by predictive coding that leverages those mechanisms.
As in other Variational Bayes RNNs, our model learns by maximizing a lower bound on the marginal likelihood of the sequential data, which is composed of two terms: the negative of the expectation of prediction errors; and the negative of the Kullback–Leibler divergence between the prior and the approximate posterior distributions. The model introduces a weighting parameter, the meta-prior, to balance the optimization pressure placed on those two terms. 
We test the model on two datasets with probabilistic structures and show that with high values of the meta-prior the network develops deterministic chaos through which the data's randomness is imitated. For low values, the model behaves as a random process. The network performs best on intermediate values, and is able to capture the latent probabilistic structure with good generalization. Analyzing the meta-prior’s impact on the network allows to precisely study the theoretical value and practical benefits of incorporating stochastic dynamics in our model. We demonstrate better prediction performance on a robot imitation task with our model using error regression compared to a standard variational Bayes model lacking such a procedure.
\clearpage



\section{Introduction}
Predictive coding has attracted considerable attention in cognitive neuroscience, as a neuroscientific model unifying possible neuronal mechanisms of prediction, recognition, and learning (\cite{rao1999predictive, lee2003hierarchical, clark2015surfing, friston2018does}). Predictive coding suggests that first, agents predict future perception through a top-down internal process. Then, prediction errors are generated by comparing the actual perception and the predicted ones. These errors are propagated through a bottom-up process to update agents' internal states such that the error is minimized and the actual perceptual inputs are recognized. Learning may then be achieved by optimizing the internal model.

Tani and colleagues \parencite{tani1999learning, tani2003self,  tani2004self} investigated neural networks which may be considered analogous to the predictive-coding framework, especially for learning temporal patterns in robotic experiments. They used recurrent neural networks (RNNs) \parencite{elman1990finding, jordan1997serial, hochreiter1997long} since RNNs are capable of learning long-term dependencies in temporal patterns. However, their predictive ability is limited in real-world applications where high uncertainty is involved. This limitation is mainly due to conventional RNNs being able to only predict perceptual inputs deterministically.

To help solve this, Murata and colleagues \parencite{murata2013learning, murata2017learning} proposed a stochastic RNN. In this RNN, the uncertainty in data is estimated by the mean and variance of a Gaussian distribution in the output layer via learning. The hidden layers stayed deterministic, however, because there was no known way to do backpropagation through random variables. This therefore limited the network from fully extracting the probabilistic structures of the target data during learning.

To work around this limitation, \textcite{kingma2013auto}, in their work on variational Bayes autoencoders (VAEs), developed a technique called the \emph{reparameterization trick} which allows to backpropoagate errors through hidden layers with random variables, thus allowing for internal stochasticity in neural networks.

\textcite{kingma2013auto} used this method in an autoencoder in order to approximate a posterior distribution of latent variables. The variational Bayes (VB) approach optimizes the network by maximizing a variational lower bound on the marginal likelihood of the data, and the prior distribution is sampled from a standard normal Gaussian. This lower bound is composed of two terms: the negative of the prediction error and the negative of the Kullback-Leibler (KL) divergence between the approximate posterior and prior distributions.

Various RNNs have been proposed based on the VAE. The first variational Bayes RNNs proposed sampling the prior distribution from a standard normal Gaussian at each timestep \parencite{fabius2014variational, bayer2014learning}. Later, \textcite{chung2015recurrent} proposed a VAE RNN called the Variational RNN (VRNN). The VRNN used a conditional prior distribution derived from the state variables of an RNN to account for temporal dependencies within the data. Since then, there have been various attempts to modify the approximate posterior of the VRNN. Some recent studies proposed approximate posteriors that had more similar structures to the true posterior by considering future dependencies on sequential data by using two RNNs---one forward and one backward \parencite{fraccaro2016sequential, goyal2017z, shabanian2017variational}. Another issue targets VRNN-based models: they have a tendency to ignore the stochasticity introduced by their random variables, and to rely on deterministic states only. To remedy this, there have been several attempts to 'force' the latent variables to learn meaningful information in the approximate posteriors (\cite{bowman2015generating, karl2016deep, goyal2017z}).

The current paper proposes a novel network model referred to as the predictive-coding-inspired variational RNN (PV-RNN) that integrates ideas from recent variational RNNs and predictive coding. In this model, the prior distribution is computed using conditional parameterization similar to \textcite{chung2015recurrent} whereas the posterior is approximated using a new adaptive vector $\bm{A}$ which forces the latent variables to represent meaningful information. This new vector also provides the approximate posterior with the future dependency information via backpropagation through time (BPTT) \parencite{werbos1974beyond, rumelhart1985learning} \emph{without} a backward RNN. All model variables and $\bm{A}$ are optimized by maximizing a variational lower bound on the marginal likelihood of the data.

Our model also incorporates a process inspired by the predictive coding framework, \emph{error regression}, which is used online during testing in our experiments after learning is finished. During error regression, the model constantly makes predictions, and the resulting prediction errors are backpropagated up the network hierarchy to update the internal states $\bm{A}$ of the model, in order to maximize both negative terms of the lower bound. 

Many studies have assumed that the brain may use predictive coding to minimize a free energy or maximize a lower bound on surprise \parencite{friston2005theory,friston2010free, hohwy2013predictive, clark2015surfing}. By incorporating features inspired by predictive coding principles, our model may be considered to be more consistent with the ideas of computational neuroscience than other VAE-based models. While most models propagate inputs through the network during the forward computation, our model only propagates prediction errors through backpropagation through time (BPTT).

One important motivation in the current study is to clarify how uncertainty or probabilistic structure hidden in fluctuating temporal patterns can be learned and then internally represented in the latent variables of an RNN. Importantly, randomness hidden in sequences can be accounted either by deterministic chaos or a stochastic process. Therefore, if we consider that we may observe sensory data with only finite resolution (\cite{crutchfield1992semantics}), then if the original dynamics are chaotic, the symbolic dynamics observed through Markov partitioning involved with the coarse-graining may be ergodic, generating stochasticity (\cite{sinai1972gibbs}). Inversely, if a deterministic RNN acts as a generative model to reconstruct such stochastic sequences through learning, the RNN may do so by embedding the sequences into internal, deterministic chaos by leveraging initial sensitivity (\cite{tani1995embedding}). An interesting question, however, is: if a generative model contains an adaptive mechanism to estimate first-order statistics---as in our proposed model, and other variational Bayes RNNs---how may the components of deterministic and stochastic dynamics be used to account for observed stochasticity in the model's output?

To examine this question, we introduce a variable, the \emph{meta-prior}, that weights the minimization of the divergence between the posterior and the prior against that of the prediction error in the computation of the variational lower bound. We investigate how the meta-prior influences development of different types of information processing in the model. First, we conduct a simulation experiment using a simple probabilistic finite state machine (PFSM) and observe how different settings of the meta-prior affect representation of uncertainty in the latent state of the model. Next, we examine how different representations of latent states in the model can lead to development of purely deterministic dynamics, random processes, or something in-between these two extremes. In particular, we examine how generalization capabilities correlate with such differences.  

Next, we consider a more complex setup, where the data embeds multi-timescale information and the network features multiple layers with each its own time constant. This allows the model to deal with fluctuating temporal patterns that consist of sequences of hand-drawn primitives with probabilistic transitions among them. We conduct simulation experiments to examine if the multiple-layer model exhibits qualitatively the same ability as the one of the previous experiment to extract the latent probabilistic structures of such compositionally-organized sequence data.

Finally, we evaluate performance of the proposed model in a real-world setting by conducting a robotic experiment. On a task where a robot learns to imitate another, imitation performance is compared between PV-RNN with error regression for posterior inference, and VRNN, which uses a variational autoencoder. This experiment aims to evaluate our hypothesis that the posterior inference calculated through error regression provides better estimates than an autoencoder-based model.
\clearpage


\section{Model}

\newcommand*{\Px}[1][]{\ensuremath{P_{\theta_X}(\bm{X}_{t}~|~{#1{\bm{d}}_{t}, \bm{Z}_{t}})}}
\newcommand*{\Pz}[1][]{\ensuremath{P_{\theta_Z}(\bm{Z}_{t}~|~{#1{\bm{d}}_{t-1}})}}
\newcommand*{\Pd}{\ensuremath{P_{\theta_d}(\bm{d}_{t}~|~\bm{d}_{t-1}, \bm{Z}_{t})}}

Let's now describe in detail the generative and inference models, as well as the learning procedure. The generative model produces predictions based on the latent state of the network. Conversely, the inference model, given an observation, estimates what should be the latent state in order to produce the observation. The learning process concerns itself with discovering good values for the learnable variables of both the generative and inference models. 

\begin{figure}[t!]
\hfill
\begin{center}
\includegraphics[width=5in]{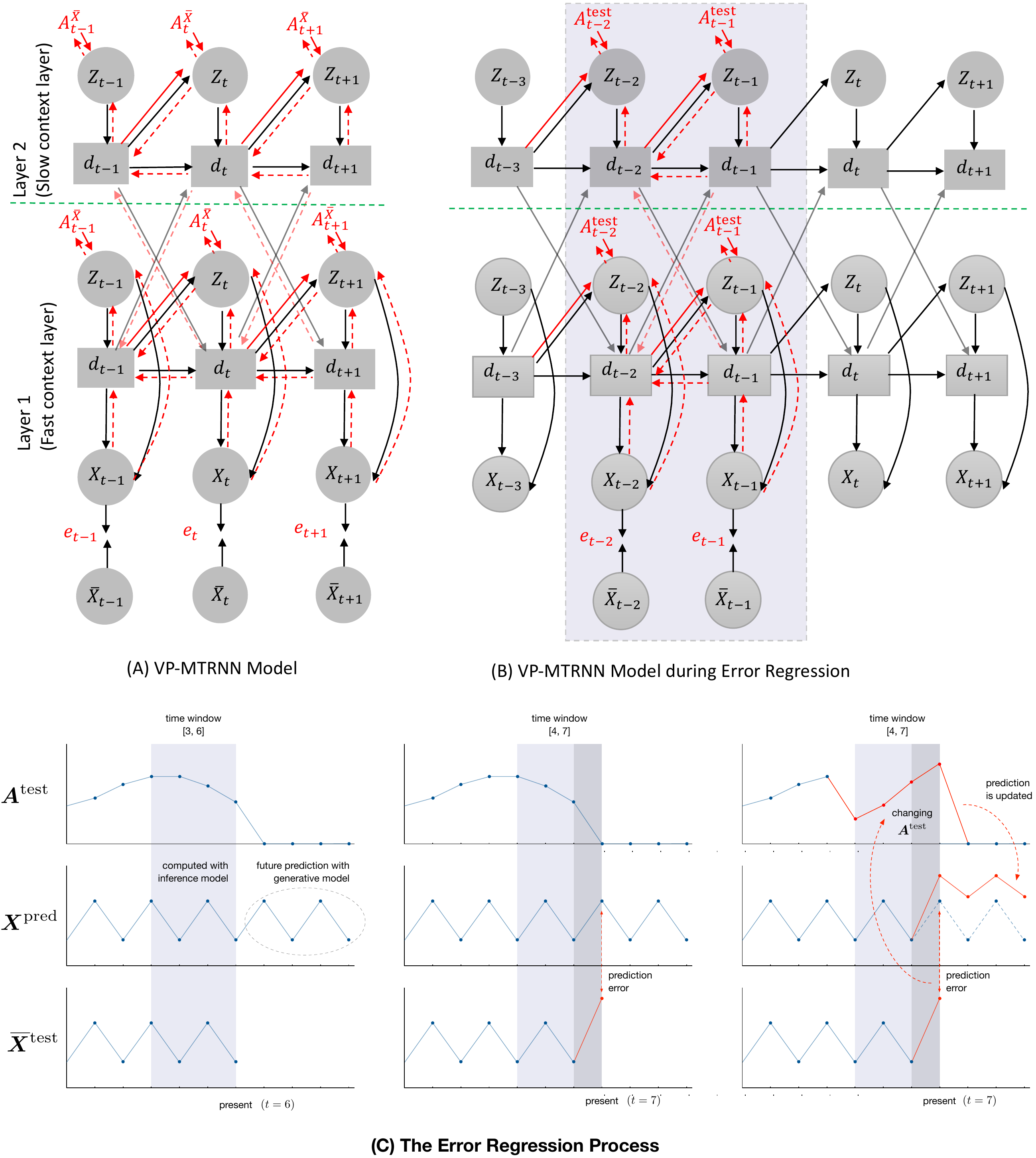}
\end{center}
\caption{(A) the generative and inference models of PV-RNN, in an MTRNN setting (B) the error regression graph during tests and (C) the error regression process. In (A) and (B), black lines represent the generative model and red lines show the inference model, with solid red ones showing the feed-forward computations of the inference model and dashed red lines showing the BPTT that is used to update $\bm{A}^{\widebar{\bm{X}}}$ in (A) and $\bm{A}^{\textrm{test}}$ in (B). The gray area in (B) represents a 2-step temporal window of the immediate past in which $\bm{A}_{t-2:t-1}^{\textrm{test}}$ is modified to maximize the lower bound. (C) illustrates the error regression process. At $t=6$, predictions are generated (left) after observing $\widebar{\bm{X}}_{1:6}^{\textrm{test}}$. The 3-timestep time-window is slid by one timestep to $[4, 7]$ (middle; now, $t=7$), and an error is observed between the prediction $\bm{X}_{4:7}^{\textrm{pred}}$ and the target value $\widebar{\bm{X}}_{4:7}^{\textrm{test}}$. The lower bound is computed and backpropagation is performed; $\bm{A}_{4:7}^{\textrm{test}}$ is then optimized and the prediction $\bm{X}_{4:7}^{\textrm{pred}}$ is updated (right). This backpropagation/optimization/prediction cycle can be repeated multiple times before moving on to the $[5, 8]$ time-window.}
\label{Fig1}
\end{figure}

\subsection{Generative Model}
As with many published Variational Bayes models, the generative model $P_\theta$ of PV-RNN is an RNN with stochastic latent variables. Here, $\theta$ denotes the learnable variables of the generative model, which is illustrated in Figure~\ref{Fig1}(A) by black lines. The variables $\theta$ are distributed among the components $\bm{X}, \bm{Z}, \bm{d}$ of the generative model, as $\theta_X, \theta_Z, \theta_d$. $\bm{Z}$ and $\bm{d}$ are the stochastic and deterministic latent states, respectively, and $\bm{X}$ is the generated prediction. For a prediction $\bm{X}_{1:T} = (\bm{X}_{1}, \bm{X}_{2}, ..., \bm{X}_{T})$, the generative model factorizes as:

\alignedequation{eq:gen_model}{
    P_\theta(\bm{X}_{1:T}, \bm{Z}_{1:T}, \bm{d}_{1:T} | \bm{Z}_{0}, \bm{d}_{0}) &= P_{\theta_X}(\bm{X}_{1:T}~|~\bm{d}_{1:T}, \bm{Z}_{1:T}) P_{\theta_Z}(\bm{Z}_{1:T}~|~\bm{d}_{1:T}, \bm{Z}_{0}) P_{\theta_d}(\bm{d}_{1:T}~|~\bm{Z}_{1:T}, \bm{d}_{0}) \\
    &= \prod_{t=1}^{T} \Px \Pz \Pd %
}

The initial values of $\bm{Z}$ and $\bm{d}$ at timestep zero, $\bm{Z}_{0}$ and $\bm{d}_{0}$, are set to $0$ in our experiments. The latent state $\bm{d}_{t}$ is recursively computed using an RNN model: 
\alignedequation{eq:f_theta_d}{
    \bm{d}_{t} = f_{\theta_d}(\bm{d}_{t-1}, \bm{Z}_{t})
}

In this paper, we use a Multiple Timescale Recurrent Neural Network (MTRNN) (\cite{yamashita2008emergence}) as $f_{\theta_d}$ but any type of RNN, such as LSTMs or GRUs could be used instead. MTRNNs are a type of RNNs composed of several hierarchical layers, with each layer using a different time constant. The internal dynamic of an MTRNN model is computed as:

\alignedequation{eq:mtrnn}{
    \bm{h}_{t}^{k} &= (1 - \frac{1}{\tau_k})\bm{h}_{t-1}^{k} + \frac{1}{\tau_k}(\bm{W}^{kk}_{dd} \bm{d}_{t-1}^k + \bm{W}^{kk}_{dz} \bm{Z}_{t}^k + \bm{W}^{kk+1}_{dd} \bm{d}_{t-1}^{k+1} + \bm{W}^{kk-1}_{dd} \bm{d}_{t-1}^{k-1})\\
    \bm{d}_{t}^k &= \tanh(\bm{h}^k_{t})
}
where $\bm{h}_{t}^{k}$ is the vector of the internal state values of the $k_{th}$ context layer at time $t$, $\bm{W}^{kk}_{dd}$ is the matrix of the connectivity weights from the $\bm{d}$ units in the $k_{th}$ context layer to itself, $\bm{W}^{kk}_{dz}$ the connectivity weights from $\bm{Z}$ to $\bm{d}$ in layer $k$, $\bm{W}^{kk+1}_{dd}$ is the matrix of the connectivity weights from the $\bm{d}$ units of the ${k+1}_{th}$ context layer to the ones in the ${k}_{th}$ layer, and similarly $\bm{W}^{kk-1}_{dd}$ is the matrix for the one coming from layer $k-1$, and $\tau$ is the time constant. Bias terms are not shown in Equation \ref{eq:mtrnn} for clarity. In this paper, we will consider networks with no more than three layers. Also, as is common with MTRNNs, the lower layer will have a faster time constant than the higher layer. In that context, we will refer to the lowest layer, with the fastest time constant, as the \emph{fast layer}, and symmetrically, to the highest layer, with the slowest time constant, as the \emph{slow layer}. The slow (highest) layer does not have any layer above it, and so, obviously, in Equation \ref{eq:mtrnn} the term $\bm{W}^{kk+1}_{dd} \bm{d}_{t-1}^{k+1}$ is removed. The same thing applies for the fast (lowest) layer and the term $\bm{W}^{kk-1}_{dd} \bm{d}_{t-1}^{k-1}$. Figure~\ref{Fig1}(A) shows the PV-RNN model implemented with a two-layer MTRNN. We extended the original MTRNN model (\cite{yamashita2008emergence}) by adding stochastic units $\bm{Z}$ to each layer. Each layer communicates only with the layer above and the one below to create a hierarchical structure.

Finally, the prior distribution $P_{\theta_Z}(\bm{Z}_{t}~|~\bm{d}_{t-1})$ is a Gaussian with diagonal covariance matrix which depends on $\bm{d}_{t-1}$. Priors depending on the previous state were used in (\cite{chung2015recurrent}) and it outperformed the independent standard Gaussian prior used in STORN (\cite{bayer2014learning}). 
\alignedequation{eq:prior_distrib}{
    \Pz = \mathcal{N}(\bm{Z}_{t}; \bm{\mu}_t^{(p)}, \bm{\sigma}_t^{(p)}) ~~~ \mathrm{where} ~~~ [\bm{\mu}_t^{(p)}, \log{\bm{\sigma}_t^{(p)}}]= f^{(p)}_{\theta_Z}(\bm{d}_{t-1})
}
where $f^{(p)}_{\theta_Z}$ denotes a one layer feed-forward neural network, and $\bm{\mu}_t^{(p)}$ and $\bm{\sigma}_t^{(p)}$ are the mean and standard deviation of $\bm{Z}_{t}$. We use the reparameterization trick (\cite{kingma2013auto}) such that the latent value $\bm{Z}$ in both posterior and prior are reparameterized as $\bm{Z}_{ }= \bm{\mu} + \bm{\sigma} \ast \bm{\varepsilon}$, where $\bm{\varepsilon}$ is sampled from $\mathcal{N}(0, I)$. In this study, $P_{\theta_X}(\bm{X}_{t}~|~\bm{d}_{t}^1, \bm{Z}_{t}^1)$ is obtained by a one layer feed-forward model $f^{(x)}_{\theta_X}$. 

One peculiar detail about the generative model is that it does not accept any external inputs. Indeed, the generative model, unlike many other variational Bayes RNN models, generates sequences based on the latent state exclusively. Rather than using external inputs, the PV-RNN model propagates the errors between the predictions and the observations via backpropagation through time. To understand this clearly, we need to explain the inference model first.

\subsection{Inference Model}\label{Inference Model}
Based on the generative model, the true posterior distribution of $\bm{Z}_{t}$ depends on $\bm{X}_{t:T}$, which can be verified using d-separation (\cite{geiger1990identifying}). Computing the true posterior is intractable, so an inference model is designed to compute an approximate posterior.

To compute $\bm{Z}_{t}$, the network considers the deterministic state of the network during the previous timestep, $\bm{d}_{t-1}$. In all other variational Bayes RNNs, $\bm{d}$ units are fed training patterns directly, but in our case, we removed those inputs to force $\bm{d}$ not to ignore $\bm{Z}$. We need another method to feed the network with information specific to the current pattern. To that end, for a training sequence of $T$ timesteps $\widebar{\bm{X}}_{1:T}$, we introduce the adaptive vectors $\bm{A}_{1:T}^{\widebar{\bm{X}}}$. For each timestep $\widebar{\bm{X}}_{t}$ of $\widebar{\bm{X}}$, we have a corresponding vector $\bm{A}_{t}^{\widebar{\bm{X}}}$. This vector is specific to the sequence $\widebar{\bm{X}}$. In other words, the model is going to have $T\times N_{\widebar{\bm{X}}}$ adaptive vectors like this, with $N_{\widebar{\bm{X}}}$ the number of training sequences.

Each $\bm{A}_{t}^{\widebar{\bm{X}}}$ is going to be adapted through BPTT, and the changes made through BPTT will depend on the prediction errors between $\bm{X}$ and $\widebar{\bm{X}}$ from $T$ to $t$, $\bm{e}_{t:T}$. Naturally, the other learning variables of the network $\theta_X, \theta_Z, \theta_d$, and $\phi$ (see next equation for $\phi$) will also be affected during BPTT by the information contained in $\bm{e}_{1:T}$. But those variables are trained on all training patterns. Only $\bm{A}^{\widebar{\bm{X}}}$ will be specifically trained on the prediction errors relative to $\widebar{\bm{X}}$. As such, $\bm{A}_{t}^{\widebar{\bm{X}}}$ is able to specifically capture information about the future timesteps $\widebar{\bm{X}}_{t:T}$ of the training sample and their existing dependencies with the current timestep $t$. Then, during inference, $\bm{A}_{t}^{\widebar{\bm{X}}}$ and $\bm{d}_{t-1}$ are combined to compute the mean and standard deviation $\bm{\mu}_t^{(q)}$ and $\bm{\sigma}_t^{(q)}$ that define the distribution from which $\bm{Z}_{t}$ will be drawn. It is to this mechanism that we will be referring to the rest of the article when we claim that we do not directly feed the external inputs to the network during the forward computation but instead, the prediction errors, and thus information about future observations, are propagated through the network via BPTT. The idea to convey information about future observations is also present in Variational Bi-LSTMs (\cite{fraccaro2016sequential, goyal2017z, shabanian2017variational}) although they use a backward RNN for this purpose, and therefore a feedforward mechanism, rather than backpropagation as we do here. 

The approximate posterior is obtained as:
\alignedequation{eq:posterior_distrib}{
    q_{\phi}(\bm{Z}_{t}~|~\bm{d}_{t-1}, \bm{e}_{t:T}) = \mathcal{N}(\bm{Z}_{t}; \bm{\mu}_t^{(q)}, \bm{\sigma}_t^{(q)}) ~~~ \mathrm{where} ~~~ [\bm{\mu}_t^{(q)}, \log{\bm{\sigma}_t^{(q)}}]= f^{(q)}_{\phi}(\bm{d}_{t-1}, \bm{A}_{t}^{\widebar{\bm{X}}})
}
where $f^{(q)}_{\phi}$ is a one-layer feed-forward network, and $\phi$ denotes the posterior parameters. Detailed computations of $\bm{A}_{t}^{\widebar{\bm{X}}}$ in $\bm{\mu}_t^{(q)}$ and $\log{\bm{\sigma}_t^{(q)}}$ are given in Appendix~\ref{Posterior}.

Using  $\bm{A}_{t}^{\widebar{\bm{X}}}$ vectors in our model presents another advantage. In all other variational Bayes RNNs, $\bm{d}$ units are fed the training patterns directly, and the network can solely rely on $\bm{d}$ to regenerate the training pattern, ignoring $\bm{Z}$ during learning and making it largely irrelevant in the computation (\cite{bowman2015generating, karl2016deep, kingma2016improved, chen2016variational, zhao2017learning, goyal2017z}). In our proposed model, if $\bm{d}$ ignores $\bm{Z}$, then it has no access to pattern specific information. This is one reason why $\bm{A}_{t}^{\widebar{\bm{X}}}$ vectors target $\bm{Z}_{t}$ and not $\bm{d}_{t}$, to avoid ignoring $\bm{Z}_{t}$ during training. On top of that, in our implementation, $\bm{Z}_{t}$ has a 10 times smaller dimension than $\bm{d}_{t}$, making it more efficient for $\bm{A}_{t}^{\widebar{\bm{X}}}$ to target $\bm{Z}_{t}$ than $\bm{d}_{t}$. One might wonder if rather than introducing new latent vectors $\bm{A}_{t}^{\widebar{\bm{X}}}$, we might have directly replaced $\bm{Z}_{t}$ by $\bm{A}_{t}^{\widebar{\bm{X}}}$ during the posterior computation. We did not do this for two reasons. First, we wanted to keep the structure of the prior and posterior as close as possible. Second, we assumed that providing the information about the past $\bm{d}_{t-1}$ to the posterior computation of $\bm{Z}_{t}$ would be beneficial in some context. This assumption is tested in Appendix \ref{appendix:exp2DiffInfModel}.

\subsection{Learning Process}

\newcommand*{\evidence}{\ensuremath{P_\theta(\bm{X}_{1:T} | \bm{Z}_{0}, \bm{d}_{0})}}
\newcommand*{\qphi}{q_{\phi}(\bm{Z}_t | \tilde{\bm{d}}_{t-1}, \bm{e}_{t:T})}
\newcommand*{\PzqphiPx}{\frac{\Pz[\tilde]}{\qphi} \Px[\tilde]}

To learn the variables $\theta$ and $\phi$ of the generative and inference models, we need to define a loss function. For variational Bayes neural networks, it has been shown that models' variables can be jointly learned by maximizing a lower bound on the marginal likelihood of training data (\cite{kingma2013auto, bayer2014learning, chung2015recurrent, fraccaro2016sequential, goyal2017z}). We maximize a lower bound because maximizing the marginal likelihood directly is intractable. Let's derive the lower bound now.

Based on Equation \ref{eq:gen_model}, the marginal likelihood or evidence can be expressed as 
\alignedequation{eq:marginal_likelihood}{    
    \evidence = \iint \prod_{t=1}^{T} [\Px \Pz \Pd] d\bm{Z}_{1:T}~d\bm{d}_{1:T}
}

Given $\bm{d}_{t-1}$ and $\bm{Z}_{t}$, the value of $\bm{d}_{t}$ is deterministic. Therefore, if we denote the \emph{value} of the variable $\bm{d}_{t}$ as $\tilde{\bm{d}}_{t}$ (equal to $f_{\theta_d}(\bm{d}_{t-1}, \bm{Z}_{t})$, as per Equation \ref{eq:f_theta_d}), \Pd{} is a Dirac distribution centered on $\tilde{\bm{d}}_{t}$. By replacing \Pd{} by the Dirac delta function $\delta(\bm{d}_t - \tilde{\bm{d}_t})$ in Equation \ref{eq:marginal_likelihood}, we can remove the integral over $\bm{d}$:
\alignedequation{eq:single_int}{    
    \evidence &= \int \prod_{t=1}^{T} \left[\Px[\tilde] \Pz[\tilde]\right] d\bm{Z}_{1:T}
}
If we factorize the integral over time and take the logarithm of the marginal likelihood, we will have:
\alignedequation{eq:log_sum}{    
    \log{\evidence} &= \log \prod_{t=1}^{T} \left[\int \Px[\tilde] \Pz[\tilde] d\bm{Z}_{t}\right] \\
                    &= \sum_{t=1}^{T} \log  \left[\int \Px[\tilde] \Pz[\tilde] d\bm{Z}_{t}\right] \\
}

Let's now multiply the inside of the integral by $1 = \frac{\qphi}{\qphi}$, in order to obtain an expectation form. Also, this introduces the inference model into equations that were generative-model-only so far, allowing for the joint optimization of both models. 
\alignedequation{eq:9}{
    \log{\evidence} &= \sum_{t=1}^{T} \log \underbrace{\left[\int \qphi \PzqphiPx d\bm{Z}_{t}\right]}_{E_{\qphi}\left[\PzqphiPx\right]}
}

Since logarithm is a concave function, we can apply Jensen's inequality: $\log(E[X]) \geq E[\log(X)]$ 
\alignedequation{eq:10}{
    \log{\evidence} &= \sum_{t=1}^{T} \log \left[\int \qphi \PzqphiPx d\bm{Z}_{t}\right] \\
    &\geq \underbrace{\sum_{t=1}^{T} \int \qphi \log\left[\PzqphiPx\right] d\bm{Z}_{t}}_{L(\theta, \phi): \textrm{Variational Evidence Lower Bound}}
}
Now, the Variational Evidence Lower Bound (ELBO) $L(\theta, \phi)$ can be maximized instead of the logarithm of the marginal likelihood \evidence{} in order to optimize the learning variables of the generative model and the approximate posterior. This formula for maximizing the lower bound is equivalent to the principle of free energy minimization provided by Friston~(\cite{friston2005theory}). $L(\theta, \phi)$ can be rewritten as:
\alignedequation{eq:elbo_kldiv}{
    L(\theta, \phi) &= \sum_{t=1}^{T} (\int \qphi \log{\Px[\tilde]}d\bm{Z}_{t} - \int \qphi \log{\frac{\qphi}{\Pz[\tilde]}}d\bm{Z}_{t}) \\
                    &= \sum_{t=1}^{T} \left(E_{\qphi}[\log{\Px[\tilde]}] - KL[\qphi~||~{\Pz[\tilde]}]\right)
}
where the first term on the right-hand side is the expected log-likelihood under $\qphi$ or the negative of the expected prediction error (\cite{kingma2013auto}), and the second term is the negative Kullback-Leibler (KL) divergence between the posterior and prior distributions of the latent variables. Only the summation over time is shown in this equation, but the lower bound is also summed over the number of training samples. We divided the first term by the dimension of $\bm{X}$ and the second term by the dimension of $\bm{Z}$ during experiments. The KL divergence is computed analytically as:
\alignedequation{eq:elbo_analytical}{
    KL[\qphi~||~\Pz[\tilde]] = \log{\frac{\bm{\sigma}_t^{(p)}}{\bm{\sigma}_t^{(q)}}}+\frac{(\bm{\mu}_t^{(p)} - \bm{\mu}_t^{(q)})^2 + (\bm{\sigma}_t^{(q)})^2}{2(\bm{\sigma}_t^{(p)})^2} -\frac{1}{2}
}
which is simply the KL divergence between two Gaussian distributions. The detailed derivation of the KL divergence is in Appendix~\ref{KL}. 

The variables of the prior $\theta_{Z}$ are optimized through the KL divergence term, whereas variables of the posterior $\phi$ are optimized through both terms. We can exploit this asymmetry: by weighting the two terms differently, we can increase or decrease the explicit optimization pressure on the learning variables corresponding to the prior or the posterior. To that end, we introduce a weighting parameter, the \emph{meta-prior} $w$, in the lower bound (Equation \ref{eq:elbo_kldiv}) to regulate the strength of the KL divergence, producing:

\alignedequation{eq:13}{
    L_w(\theta, \phi) = \sum_{t=1}^{T} (E_{\qphi}[\log{\Px[\tilde]}] - w \cdot KL[\qphi~||~{\Pz[\tilde]}])
}

In the experiments, all model variables and $\bm{A}$ are optimized in order to maximize the lower bound using ADAM~(\cite{kingma2014adam}). We use the same parameter setting for the ADAM optimizer as the original paper: $\alpha$ = $0.001$, $\beta_1$ = $0.9$, and $\beta_2$ = $0.999$ in training. In both experiments, the number of latent units $\bm{Z}$ were $10$ times smaller than the number of deterministic units $\bm{d}$.

\subsection{Error Regression}

Testing the network on unseen training sequences is not straightforward: it does not accept any input during the forward computation. It does, however, propagate errors during backpropagation. So we leverage this mechanism during testing.   

While training the inference model, we created sequences of adaptive vectors $\bm{A}_{1:T}^{\widebar{\bm{X}}}$, one for each training observation. The purpose was to capture the relevant information about the training observation into $\bm{A}_{1:T}^{\widebar{\bm{X}}}$, and train the other weights of the network, $\theta$ and $\phi$, to use this information to make useful predictions. Another way to understand this is that $\bm{A}_{1:T}^{\widebar{\bm{X}}}$ are building good representations that the rest of the network, shared among all training sequences, learns to use. In that sense, the adaptive vectors $\bm{A}_{1:T}^{\widebar{\bm{X}}}$ on one side, and the weights $\theta$ and $\phi$ on the other side, are fulfilling vastly different roles. And as we will see, once the training is done, the values of $\bm{A}_{1:T}^{\widebar{\bm{X}}}$ are not needed anymore to process unseen testing sequences. 

When processing an unseen testing sequence, the weights $\theta$ and $\phi$ are fixed, and the adaptive vectors $\bm{A}_{1:T}^{\widebar{\bm{X}}}$ are unavailable. We initialize the adaptive vector $\bm{A}_{1:T^{\prime}}^{\textrm{test}}$ to zero values; we are going to optimize $\bm{A}_{1:T^{\prime}}^{\textrm{test}}$ online, during the processing of $\widebar{\bm{X}}^{\textrm{test}}$, to maximize our ability to predict it. This online optimization is done incrementally, inside a time-window of size $m$. The process is illustrated in Figure~\ref{Fig1}(C).

Using the $m$ (and for now, zero-valued) $\bm{A}_{1:m}^{\textrm{test}}$ values, we can generate $\bm{Z}_{1:m}^{\textrm{pred}}$ using the inference model $q_{\phi}$ (Equation \ref{eq:posterior_distrib}), and compute $\bm{d}_{1:m}^{\textrm{pred}}$ using Equation \ref{eq:mtrnn}. The prediction $\bm{X}_{1:m}^{\textrm{pred}}$ can also be computed using $\bm{Z}_{1:m}^{\textrm{pred}}$ and $\bm{d}_{1:m}^{\textrm{pred}}$. $\bm{X}_{1:m}^{\textrm{pred}}$ is then compared to $\widebar{\bm{X}}_{1:m}^{\textrm{test}}$, and the resulting prediction errors $\bm{e}_{1:m}$ are backpropagated through the network to update the values of $\bm{A}_{1:m}^{\textrm{test}}$. The update is done the same way the network is trained, by computing the lower bound and using BPTT, except that the variables $\theta$ and $\phi$ are fixed and are not modified. The new values of $\bm{A}_{1:m}^{\textrm{test}}$ are used to generate a new prediction $\bm{X}_{1:m}^{\textrm{pred}}$, and a new optimization cycle can occur. The number of optimization cycles of $\bm{A}_{1:m}^{\textrm{test}}$ for a given time-window can depend on reaching a given error threshold, be fixed beforehand, or, in a real-time context, depend on the available computational time. Next, the time-window is slid to $[2, m+1]$, and $\bm{A}_{2:m+1}^{\textrm{test}}$ are used to generate $\bm{X}_{2:m+1}^{\textrm{pred}}$ and are optimized. Importantly, only the part of $\bm{A}_{1:T^{\prime}}^{\textrm{test}}$ inside the time-window---here $\bm{A}_{2:m+1}^{\textrm{test}}$---is optimized. In particular, $\bm{A}_{1}^{\textrm{test}}$ is now fixed. After the optimization of $\bm{A}_{2:m+1}^{\textrm{test}}$, the time-window moves to $[3, m+2]$ and so on. 

At any point in this process, for a time-window $[t-m, t-1]$, the prediction steps outside the time-window $\bm{X}_{t}, \bm{X}_{t+1}, \bm{X}_{t+2}, ...$ can be generated by computing $\bm{Z}_{t}, \bm{Z}_{t+1}, \bm{Z}_{t+2}, ...$ using the generative model (Equation \ref{eq:prior_distrib}), which does not depend on the values of $\bm{A}_{t}^{\textrm{test}}, \bm{A}_{t+1}^{\textrm{test}}, \bm{A}_{t+2}^{\textrm{test}}, ...$ which are, at this point, zero. The predictions $\bm{X}_{t}, \bm{X}_{t+1}, \bm{X}_{t+2}, ...$ correspond to unobserved parts of the testing sequence at this point, and therefore are the model's prediction of the future. These additional predictions have no impact on the BPTT process of error regression.

Finally, let's note that the optimization can begin before the time-window is at full size, and start with time-windows $[1, 1], [1,2], ..., [1, m], [2, m+1]$ and so on. Additionally, the optimization does not need to happen at every timestep, and can for instance be triggered every 10 timesteps, with time-windows $[1, 10], [1, 20], ..., [1, m], [11, m+10], [21, m+20], ...$ (assuming here $m$ is a multiple of 10).

The error regression process was implemented in deterministic RNNs, and it was shown how it could help the generalization capability of those models \parencite{tani2003self, murata2017learning, ahmadi2017can}. This testing process through error regression bears similarities to, and is inspired by, predictive coding. Predictive coding proposes that the brain is continually making predictions about incoming sensory stimuli, and that error between the prediction and the real stimuli is propagated back up through the layers of the processing hierarchy. Those error signals are then used to update the internal state of the brain, impacting future predictions. Our network goes through similar stages during error regression: predictions are made ($\bm{X}^{\textrm{pred}}$), compared to actual observations ($\widebar{\bm{X}}^{\textrm{test}}$), and the errors ($\bm{e}_{1:m}$) are backpropagated to update the internal state of the network ($\bm{A}_{1:m}^{\textrm{test}}$). To be very clear here, our network is not a model of the brain; it does not claim to explain any existing neurological data nor make any useful predictions about animal brains. We are merely drawing inspiration from the predictive coding ideas to design new machine learning networks. In particular, in neurological models of predictive coding (\cite{rao2000predictive}), each layer makes an independent prediction and propagates the error signal to the upper processing layer only. In our network, the prediction error from the raw sensory data is backpropagated through the entire network hierarchy. This is deliberate, because we use BPTT: we adapted the ideas of predictive coding to the classical tools of recurrent neural networks.

\subsection{Related Work}

RNNs are widely used to model temporal sequences due to their ability to capture long dependencies in data. However, a deterministic RNN $\bm{d}_{t} = f(\bm{d}_{t-1}, \widebar{\bm{X}}_{t-1})$ can have problems when modeling stochastic sequences with a high signal-to-noise ratio (\cite{chung2015recurrent}). In an attempt to solve this problem, \textcite{bayer2014learning} introduced a model called STORN by inserting a set of independent latent variables (sampled from a fixed distribution) into the RNN model. Later, the VRNN model was proposed using conditional prior parameterization (\cite{chung2015recurrent}). In their model, the prior distribution is obtained using a non-linear transformation of the previous hidden state of the forward network as [$\bm{\mu}^{(p)}_t, \log(\bm{\sigma}^{(p)}_t)$] = $f^{(p)}(\bm{d}_{t-1})$. VRNN outperformed STORN by using this type of conditional prior. However in VRNN, the posterior  is inferred at each timestep without using information from future observations. A posterior inferred in such a way would be different from the true posterior. Later, this issue was considered by using two RNNs, a forward RNN and a backward one. The backward RNN was used in the posterior to transfer future observations for the current prediction (\cite{fraccaro2016sequential, goyal2017z, shabanian2017variational}). As explained, our model manages this by updating $\bm{A}^{\widebar{\bm{X}}}$ through backpropagation of the error signal.

Recent studies of generative models show that extracting a meaningful latent representation can be difficult when using a powerful decoder. The $\bm{d}$ units ignore the latent variables $\bm{Z}$ and capture most of the entropy in the data distribution (\cite{goyal2017z}). Many researchers have addressed this issue by either weakening the decoder or by annealing the KL divergence term during training (\cite{bowman2015generating, karl2016deep, kingma2016improved, chen2016variational, zhao2017learning}). In a recent attempt, the authors of $Z$-forcing also proposed an auxiliary training signal for latent variables alone, which forces the latent variables to reconstruct the state of the backward RNN (\cite{goyal2017z}). This method introduces an additional generative model and, as a result, an additional cost on the lower bound. Comparatively, our model captures information about external inputs in $\bm{A}^{\widebar{\bm{X}}}$, and the information flows to $\bm{d}$ through $\bm{Z}$, rendering the model unable to ignore its latent variables.

Therefore, in our model, those two issues, capturing future dependencies and avoiding having the network ignores its latent states are addressed with the same mechanism: the adaptive vectors $\bm{A}^{\widebar{\bm{X}}}$.

Introducing adjustable parameters in the lower bound has been studied for variational Bayes neural networks previously. KL-annealing does this \parencite{bowman2015generating}, linearly increasing the weight of the KL-divergence term from 0 to 1 during the training process, to avoid ignoring the latent variables and to improve convergence. In \textcite{higgins2017beta}, it was shown that the degree of the disentanglement in latent representations of VAE models can be improved by strengthening the importance of the KL divergence term in the lower bound. The generative factors in an image of a dog, for example, can be its color, size, and breed. Disentangling the generative factors in the model can be beneficial, as it creates latent units sensitive to the changes in a single generative factor, while being relatively invariant to changes in other factors (\cite{bengio2013representation}). Our model considers weighting the KL divergence term for a purpose different from KL annealing or disentanglement: to influence the balance between a deterministic and a stochastic representation of the data in the model.  

The current study is a continuation of our previous work (\cite{ahmadi2017bridging}) that proposed a predictive-coding variational Bayes RNN and studied the effect of weighting the KL divergence term. This model, however, was composed of only the latent variables $\bm{Z}$ and did not use the deterministic units $\bm{d}$; it also used a prior distribution with a fixed mean and standard deviation that has been shown not to be plausible for models that deal with time-series data (\cite{chung2015recurrent}). This led us to consider separating stochastic and deterministic states in the current model. It allows us to have a conditional prior.

Separating deterministic and stochastic states provides an additional advantage: it allows to have a number of $\bm{Z}$ units significantly smaller than $\bm{d}$ units. In our test, having 10 times more $\bm{d}$ units than $\bm{Z}$ units was the best balance between performance and computational time; the number of $\bm{A}$ units was always the same as the number of $\bm{Z}$ units. We use this ratio in all our experiments.

\clearpage


\section{Simulation Experiments}

We conducted simulation experiments to examine how learning in the proposed model depends on the meta-prior $w$. The first experiment investigates how the proposed model could learn to extract the latent probabilistic structure from discrete (0 or 1) data sequences generated from a simple probabilistic finite state machine (PFSM) under different settings of the meta-prior $w$. The purpose of this relatively simple experiment is to conduct a detailed analysis of the underlying mechanism of the PV-RNN when embedding the latent probabilistic structure of the data into mixtures of deterministic and stochastic dynamics. In the second experiment, a more complex situation is considered where the model is required to extract latent probabilistic structures from continuous sequence patterns (movement trajectories). For this purpose, trajectory data was generated by considering probabilistic switching of primitive movement patterns based on another predefined PFSM in which each primitive was generated with fluctuations in amplitude, velocity, and shape. Again, we examined how the performance depends on the meta-prior $w$.

\begin{figure}[h!]
\hfill
\begin{center}
\includegraphics[width=3in]{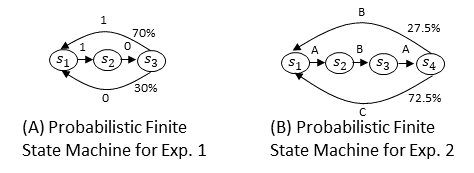}
\end{center}
\caption{The probabilistic finite state machines used to generate training patterns for PV-RNNs in (A) the first and (B) the second experiments.}
\label{AppFig1}
\end{figure}

\subsection{Experiment 1}\label{Exp1}

The PFSM shown in~Figure~\ref{AppFig1}(A) was used as the target generator. Transitions from $s_1$ to $s_2$ and $s_2$ to $s_3$ were deterministically determined with $1$ and $0$ as output, respectively. However, the transitions from $s_3$ to $s_1$ were randomly sampled with $30\%$ and $70\%$ probabilities for output $0$ and $1$, respectively. Ten target sequence patterns, of $24$ timesteps each, were generated and provided to the PV-RNN as training data. Each model had only one context layer consisting of 10 $\bm{d}$ units and a single $\bm{Z}$ unit. The time constant $\tau$ for all $\bm{d}$ units was set to $2.0$. The output of the network, $\bm{X}_{1:T}$, was discretized during testing, with outputs less than $0.5$ assigned to $0$ and the ones equal to or larger than $0.5$ assigned to $1$. Finding an adequate range of $w$ at a beginning of an experiment depends on the network parameter settings, the dataset, and the task. For this experiment, the most interesting behavior was observed in the range $[0.0001, 0.1]$. For $w$ set to larger values such as $0.5$ and $1.0$, the networks showed the same qualitative behavior to the network with $w$ set to $0.1$. Training was conducted on seven models with the different meta-prior $w$ set to $0.1$, $0.05$, $0.025$, $0.015$, $0.01$, $0.001$, and $0.0001$, respectively. In this experiment, we used an MTRNN to be consistent with our other experiments. However, it is possible to do this experiment with a simple RNN as well. Similar results were obtained by using a simple RNN and are shown in  Appendix \ref{appendix:exp1simpleRNN}.

\begin{figure}[t!]
\hfill
\begin{center}
\includegraphics[width=5in]{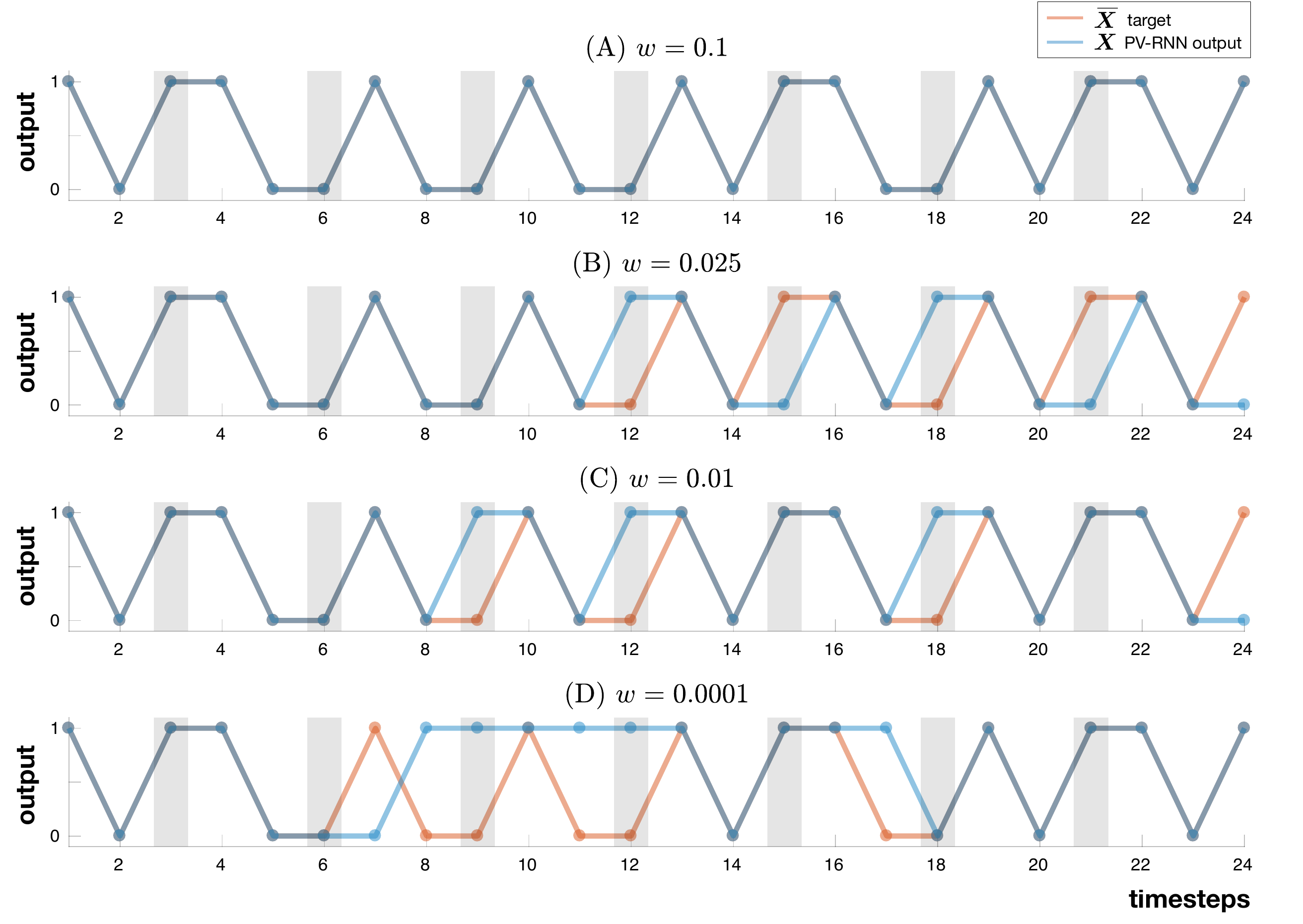}
\end{center}
\caption{Larger values of the meta-prior translate into better reconstruction of the training patterns. The four graphs show a training pattern (in orange) and its reconstruction by PV-RNN (in blue), for different values of $w$. Overlapping sections appear dark gray. For $w = 0.1$, the target sequence is completely regenerated. When $w$ is equal to $0.025$ and $0.01$, all deterministic steps are correctly reproduced, but regenerated patterns begin to diverge at the 11th and 8th step, respectively. When $w$ is set to $0.0001$, even the deterministic transition rules fail to be reproduced, and the signals diverge at the 6th timestep.}
\label{Fig2}
\end{figure}

After training for 500,000 epochs, given a training sequence $\widebar{\bm{X}}$, the learned value of $\bm{A}_{1}^{\widebar{\bm{X}}}$ is fed to the network, generating $\bm{Z}_{1}$ via Equation~\ref{eq:posterior_distrib}. Then the remaining latent states $\bm{Z}_{2:T}$ and the output $\bm{X}_{1:T}$ are generated from the generative model (Equation~\ref{eq:prior_distrib}). The purpose is to study if providing $\bm{A}_{1}^{\widebar{\bm{X}}}$ is enough for the trained network to regenerate $\widebar{\bm{X}}$ accurately. We refer to this procedure as \emph{target regeneration}.

Figure~\ref{Fig2} compares one target sequence pattern and its corresponding regeneration by the PV-RNN model trained with different values of the meta-prior. For large values of $w$ the network reproduced the training pattern accurately. As the value of $w$ decreases, divergences appear earlier and earlier, and for low values even the deterministic steps show errors.

For a given reconstruction, one can compute the diverging step as the time $t$ of the first difference between the target and the reconstruction. If both target and reconstruction are identical, the diverging step is equal to the length of the reconstruction. For each training pattern, we compute the diverging step 10 times, and compute the mean of all results to obtain the average diverging step (ADS) over the training dataset.

To characterize the deterministic nature of the network behavior, we compute the variance of the divergence (VD), which shows diversity among sequence patterns regenerated from the same value of $\bm{A}_{1}^{\widebar{\bm{X}}}$. For a given value of  $\bm{A}_{1}^{\widebar{\bm{X}}}$, we ran the regeneration 50 times, and computed the mean variance (across all 50 runs and all timesteps) of the generated $\bm{X}$ before discretization.

ADS and VD for different values of $w$ are shown on Table~\ref{tab1}. ADS decreases while VD increases as $w$ decreases. For $w = 0.1$, VD is near zero, the network reproduces the same pattern with little variation; the behavior developed can be regarded as deterministic. On the other hand, the relatively high value of VD for $w=0.0001$ points to highly stochastic dynamics.

\begin{table}[t!]
    \caption{The Average Diverging Step (ADS) and Variance of Divergence (VD) measure point to better reconstruction performance when $w$ is high. However, taking into account the KL divergence between the probabilistic distribution of the generated pattern $P(\bm{X}_{t:t+11})$ and the one of the training data $P(\widebar{\bm{X}}_{t:t+11})$ paints another picture: the network best captures the probabilistic structure of the data for an average value of $w$.}
    \centering  
    \begin{adjustbox}{max width=\textwidth}
    \begin{tabular}{c| c c c c c c c}    
    \cline{1-3} 
    \toprule
           \multicolumn{1}{}  & &\multicolumn{7}{c}{Meta-Prior $w$} \\ 
        
           \multicolumn{1}{}  &  & $0.1$   & $0.05$  & $0.025$ & $0.015$ & $0.01$ & $0.001$ & $0.0001$\\  
           
    \hline
            Average Diverging Step (ADS) & \textbf{22}  & 19 & 14 & 12 & 11 & 9 & 8\\
    \hline
            Variance of Divergence (VD) & \textbf{0.00003}  & 0.00155 & 0.0480 & 0.0499 & 0.0618 & 0.134 & 0.172\\
    \hline        
            KL div. of Test Phase & 5.040  & 2.276 & \textbf{0.0684} & 0.120 & 0.148 & 1.0679 & 5.607\\
    \bottomrule
    \end{tabular}
    \end{adjustbox}
    \label{tab1}
\end{table} 

In Table~\ref{tab1}, we also examine the ability of the network to extract the latent probabilistic structure from the data by computing the KL divergence between the probability distributions of sequences of length 12 generated by the PFSM, $P(\widebar{\bm{X}}_{t:t+11})$, and the one generated by the PV-RNN, $P(\bm{X}_{t:t+11})$ (thus characterizing how similar they are). To compute the probability distribution $P(\bm{X}_{t:t+11})$, we set $\bm{A}_{1}$ randomly, and generate a sequence of 50,000 steps using the generative model. We refer to this as \emph{free generation}. We consider the distribution of the $49,989$ sequences of length 12 $\bm{X}_{t:t+11}$ present in the sequence, and compute their distribution. For the probability distribution $P(\widebar{\bm{X}}_{t:t+11})$ we concatenate the 10 training sequences in one sequence of 240 timesteps (which is a valid output sequence of the PFSM), and compute the distribution of the 229 sequences of the length 12 we could extract from it. The resulting KL divergence measures from those two distributions for all PV-RNN models show that average values of the meta-prior capture the underlying transition probability of the PFSM from the training data best.

\begin{figure}[t!]
\hfill
\begin{center}
\includegraphics[width=5in]{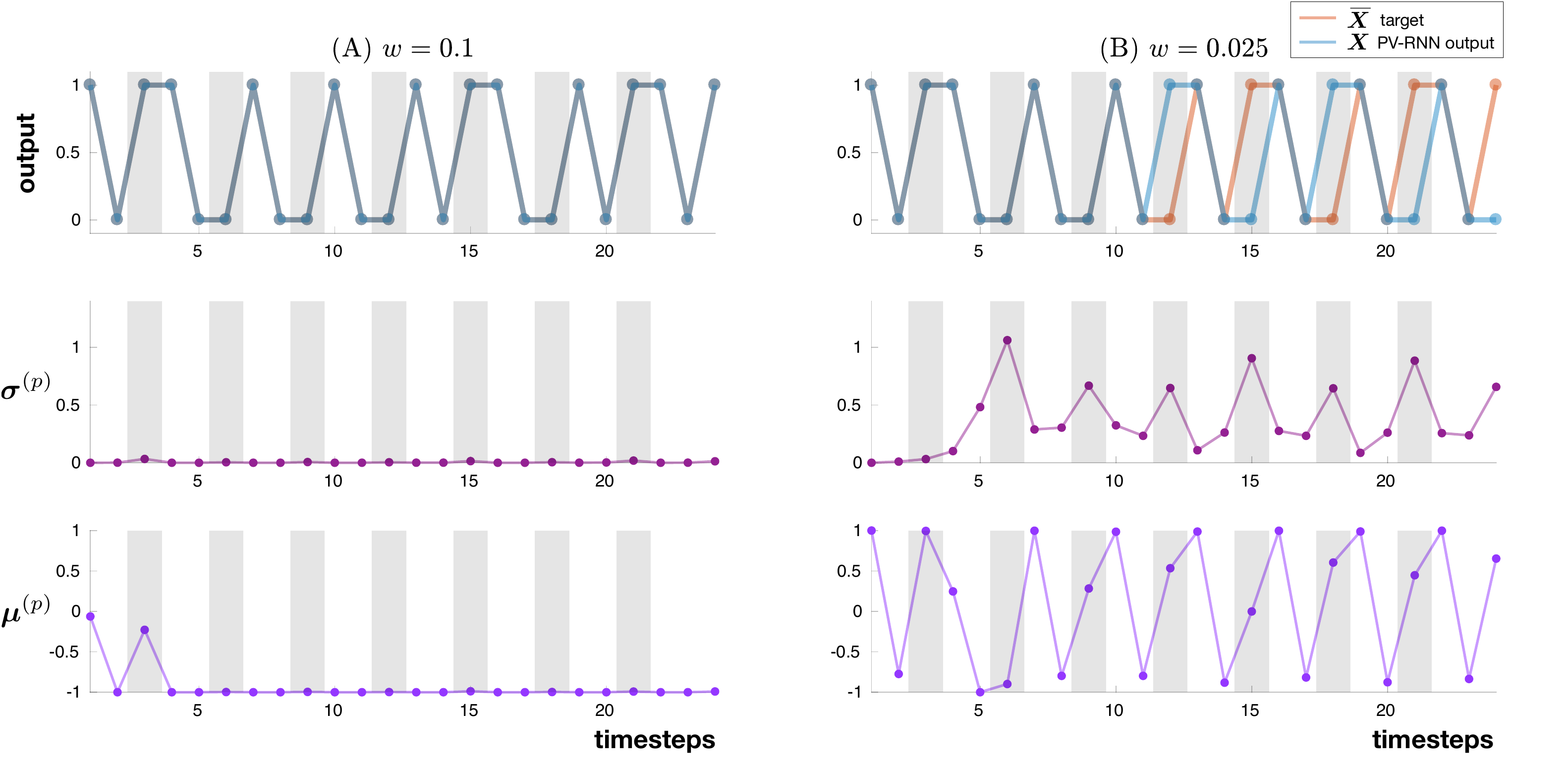}
\end{center}
\caption{ A high meta-prior forces the network into deterministic dynamics; with an average value of $w$, the probabilistic structure of the data is captured. The mean $\bm{\mu}^{(p)}$ (middle row) and variance $\bm{\sigma}^{(p)}$ (bottom row) of the latent state during regeneration of a given $\widebar{\bm{X}}$ (top row) for $w$ equal to $0.1$ and $0.025$ are shown. With $w = 0.1$, the $\bm{\sigma}^{(p)}$ is near zero: it amounts to rote learning by the network of the training pattern. $\bm{\mu}^{(p)}$, on the other hand, only varies during the first few timesteps, suggesting that the information identifying which training pattern to regenerate is transferred to the network early on, and thereafter the value of $\bm{Z}$ is disregarded. With $w = 0.025$, the variance $\bm{\sigma}^{(p)}$ is much larger overall, and significantly higher for the probabilistic states (gray bars). This suggests that PV-RNN with $w$ set to $0.025$ is capable of discriminating between deterministic and probabilistic steps in the sequence. This effect is reflected in $\bm{\mu}^{(p)}$ as well, with most deterministic states having a $\bm{\mu}^{(p)}$ close to either $1$ or $-1$, and probabilistic states mostly confined to the range $[0, 0.75]$, the asymmetry over the range possible range ($[-1, 1]$) possibly even reflecting the 70/30\% difference in transition probability.}
\label{Fig3}
\end{figure}

Figure~\ref{Fig3} displays the mean and variance of the latent state during regeneration of a target sequence for $w$ equal to $0.1$ and $0.025$ and confirms this analysis. With $w = 0.1$, the network possesses deterministic dynamics that amount to rote learning. With $w = 0.025$, the network distinguishes between deterministic and probabilistic states, and captures the probabilistic structure in its internal dynamics. Plots showing the cases of $w$ set to $0.001$ and $0.0001$ are provided in the appendix (Figure~\ref{AppFig2}). With $w=0.0001$, the value of sigma becomes high even for the deterministic case; the network does not distinguish anymore between deterministic and probabilistic states, and behaves as a random process. 

\begin{figure}[t!]
\hfill
\begin{center}
\includegraphics[width=5in]{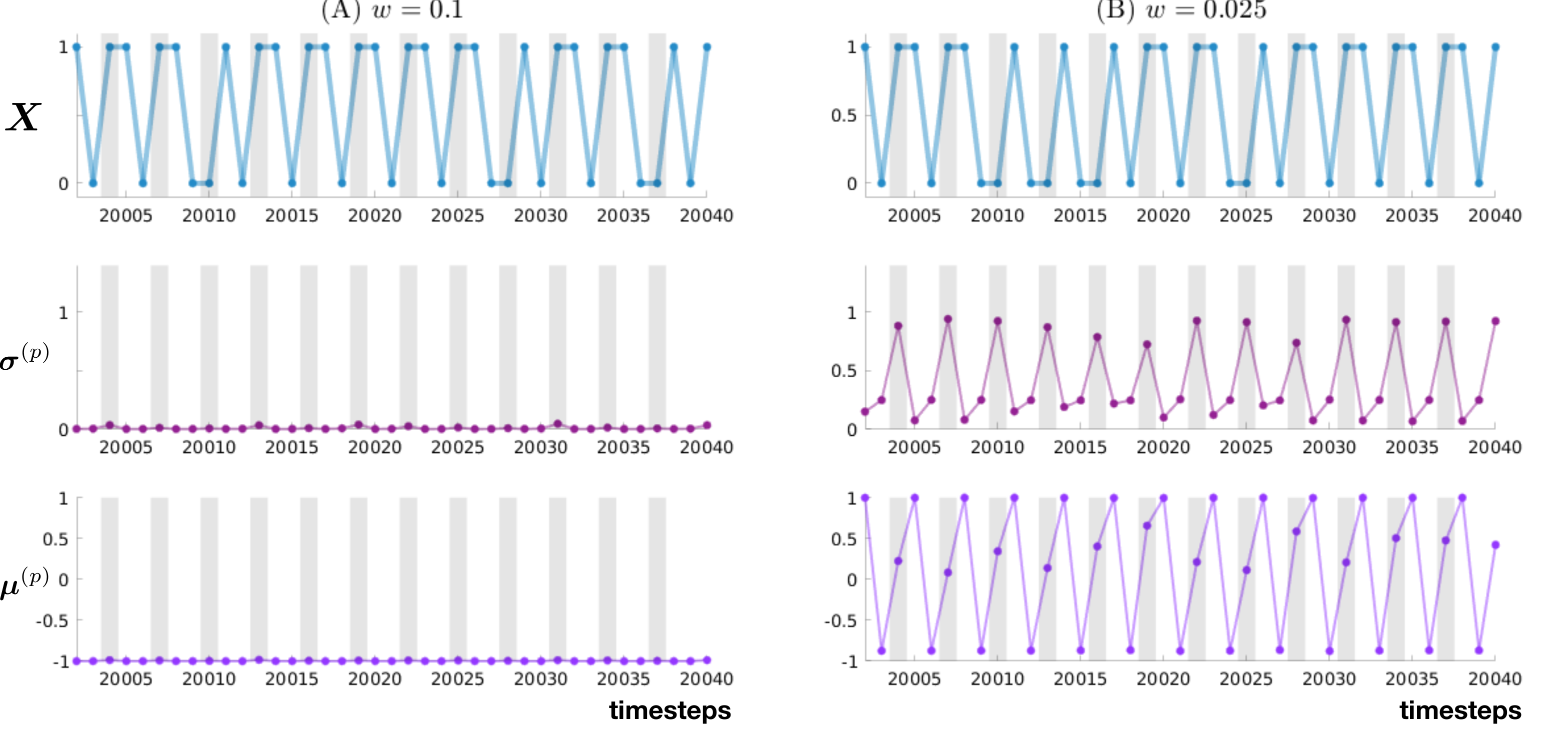}
\end{center}
\caption{ The generated output, the mean $\bm{\mu}^{(p)}$, and the standard deviation $\bm{\sigma}^{(p)}$ from timesteps $20,002$ to $20,040$ of two PV-RNNs trained with the meta-prior $w$ set to $0.1$ (A) and $0.025$ (B). Gray bars show the timesteps corresponding to uncertain states.}
\label{Fig4}
\end{figure}

Figure~\ref{Fig4} illustrates the generated output, the mean $\bm{\mu}^{(p)}$, and standard deviation $\bm{\sigma}^{(p)}$ of the $\bm{Z}$ unit for PV-RNNs trained with $w$ equal to $0.1$ and $0.025$ from timesteps $20,002$ to $20,040$. The behavior of $\bm{\mu}^{(p)}$ and $\bm{\sigma}^{(p)}$ in both cases are similar to those shown in Figure~\ref{Fig3}. In the case of $w$ set to $0.025$, the network was most successful at extracting the latent probabilistic structure from the data by detecting both the uncertain and deterministic states in the sequence. The same figure for $w$ equal to $0.001$ and $0.0001$ is shown in the appendix (Figure~\ref{AppFig3}). The transition rules defined in the PFSM were mostly broken for the case with the minimum $w$ value ($0.0001$) and frequently for the case of $w$ equal to $0.001$ as the model wrongly estimated the uncertainty as high even for the deterministic states.

We observed that the deterministic network developed with $w$ set to $0.1$ generated non-periodic output patterns. This can be roughly seen in Figure~\ref{Fig4}(A). We assumed that deterministic chaos or transient chaos developed in this learning condition. To confirm this, the Lyapunov exponents were computed using the method in \textcite{alligood1996chaos}. Interestingly, the largest Lyapunov exponent was positive. We evaluated this by generating patterns (using free generation) for 50,000 steps twice. Once as usual, and once with the random variable generating the value of the $\bm{Z}$ unit, $\bm{\varepsilon}_{1:50000}$, set to zero (so that the value of $\bm{\sigma}^{(p)}$ is irrelevant). This was done to verify that the noise generation was not impacting the value of the Lyapunov exponent. In both cases, the largest Lyapunov exponents were positive (around $0.1$). The method for computing Lyapunov exponents is described in Appendix~\ref{lya}. 

The results of the current experiment can be summarized as follows.
It was shown that different types of internal dynamics can be developed in the current model depending on the value of the meta-prior $w$ used during training on stochastic sequences. When $w$ is set to a large value ($0.1$), deterministic dynamics were generated by minimizing $\bm{\sigma}^{(p)}$ in the prior to nearly 0 for all timesteps. The deterministic aspect of the developed dynamics was further confirmed by observing that they generated the least diversity when generation was run multiple times starting from the same initial $\bm{A}_{1}$. The finding of the maximum Lyapunov exponent of the dynamics as a positive value confirmed that those dynamics developed into deterministic chaos. It was also found that the average diverging steps (ADS) became larger when $w$ was set to a larger value: each training target sequence was captured exactly for relative long timesteps, in a fashion akin to rote learning.

On the other hand, decreasing $w$ generated stochastic dynamics, even approaching the random process for low values of the meta-prior, as evidenced by the increase of diversity in sequences generated from the same latent initial state. It was, however, found that the best generalization in learning took place with $w$ set to an intermediate value. The analysis of the latent variable in this condition revealed that low values of $w$ translated into high values of $\bm{\sigma}^{(p)}$ for probabilistic \emph{and} deterministic state transitions. For intermediate values of the meta-prior, however, high values of $\bm{\sigma}^{(p)}$ were mostly observed for probabilistic state transition, indicating that the model did discriminate between the two in that case. 
To understand why this is the case, one must observe that the KL divergence term of Equation \ref{eq:13} acts as a pressure for $\bm{\sigma}^{(p)}$ to be close to $\bm{\sigma}^{(q)}$ and $\bm{\mu}^{(p)}$ to be close to $\bm{\mu}^{(q)}$: for the posterior and prior distributions to be similar to one another.

When $w$ is small, the pressure that the KL divergence term has on the backpropagation process is small to almost non-existent. Therefore, the pairs $\bm{\sigma}^{(p)}$, $\bm{\sigma}^{(q)}$ and $\bm{\mu}^{(p)}$, $\bm{\mu}^{(q)}$ are free to be uncorrelated. The other term of Equation \ref{eq:13}, the reconstruction error, puts learning pressure on $\bm{\sigma}^{(q)}$ and $\bm{\mu}^{(q)}$. Therefore, there is little learning pressure on the prior distribution, and it mostly stays close to its initialization values. In our implementation, those values are random, and therefore the network acts as a random process when the $\bm{Z}$ states are generated by the generative model.

When $w$ is high, the pressure is high for the posterior and prior distributions to be similar. Deterministic states are easier for the network to learn, and therefore, the $\bm{\sigma}^{(q)}$ and $\bm{\sigma}^{(p)}$ can both converge to small values, so as to reduce both the KL divergence term and the reconstruction error term of Equation \ref{eq:13}. Probabilistic states take longer to learn. Looking at the close-form solution of the KL-divergence term, Equation \ref{eq:elbo_analytical}, one way to reduce the KL divergence between the posterior and prior distributions is to increase $\bm{\sigma}^{(p)}$ when $\bm{\mu}^{(q)}$ and $\bm{\mu}^{(p)}$ are different. And this is the temporary solution that the network seems to be using, when looking at the evolution of $\bm{\sigma}^{(p)}$ in Figure \ref{AppFig5}. Eventually, the network makes $\bm{\sigma}^{(q)}$ and $\bm{\sigma}^{(p)}$ converge to zero in order to minimize the KL divergence further.

For the network with $w$ set to an intermediate value, the pressure is less for the posterior and prior distributions to be similar. $\bm{\sigma}^{(q)}$ and $\bm{\sigma}^{(p)}$ do not converge to zero, and the network seems to stay in the intermediate solution.


\subsection{Experiment 2}\label{Exp2}

\newcommand{\wprime}{w^{\prime}}

In this experiment, the PV-RNN was required to extract latent probabilistic structures from observed continuous sequence data (movement trajectories). 48 400-timestep data sequences and one of length 6400 timesteps were generated using the PFSM depicted in Figure~\ref{AppFig1}(B), where the primitive pattern A, B, and C corresponded to a circle, a figure-eight, and a triangle, respectively. The sequences were based on human hand-drawn patterns with naturally varying amplitude, velocity, and shape. One of such sequence can be seen in the appendix (Figure~\ref{AppFig4}). 16 of the 48 400-step sequences were used to train the model, while the 32 remaining ones were reserved for testing. The details of the generation can be found in Appendix \ref{appendix:exp2data}.

For this experiment, the most interesting behavior was observed with $w$ in the range $[1.0 \times 10^{-3}, 0.01  \times 10^{-3}]$. To avoid excessive notation in the following text, let us introduce $\wprime = w \times 10^{3}$, so that when $w$ evolves in the range $[1.0 \times 10^{-3}, 0.01  \times 10^{-3}]$, $\wprime$ evolves in $[1.0, 0.01]$.

Six PV-RNN models were trained with $\wprime$ set to $1.0$, $0.5$, $0.25$, $0.15$, $0.1$, and $0.01$. Each model had three context layers consisting of $80$ $\bm{d}$ units and $8$ $\bm{Z}$ units for the fast context (FC) layer, $40$ $\bm{d}$ units and $4$ $\bm{Z}$ units for the middle context (MC) layer, and $20$ $\bm{d}$ units and $2$ $\bm{Z}$ units for the slow context (SC) layer. The time constants of FC, MC, and SC units were set to 2, 4, and 8, respectively. Training ran for $250,000$ epochs in each case. We also conducted experiments using two context layers for two PV-RNN models with $\wprime$ set to $0.25$. The results can be seen in Appendix \ref{appendix:exp2layers}.

\begin{figure}[t!]
\hfill
\begin{center}
\includegraphics[width=5in]{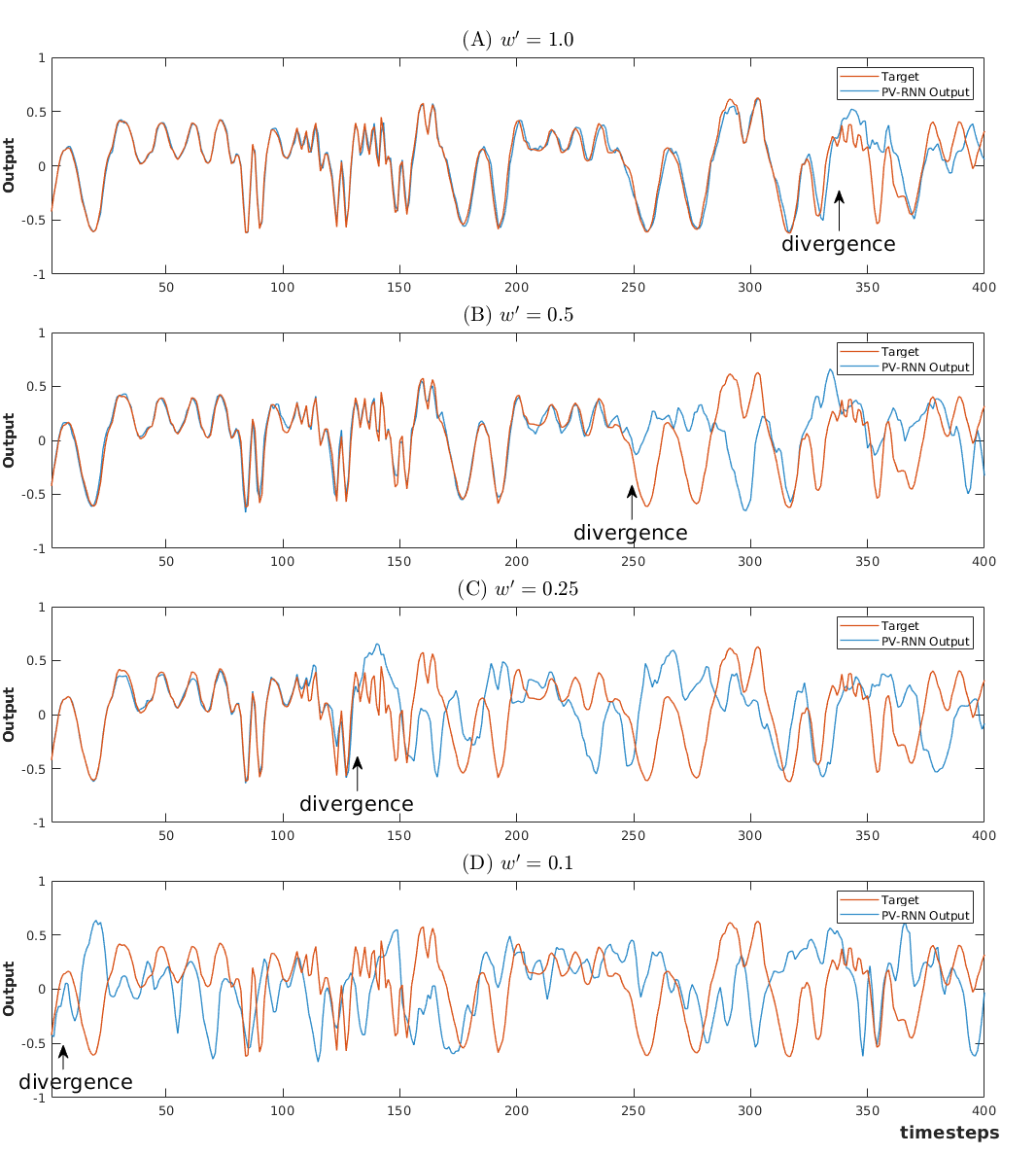}
\end{center}
\caption{ PV-RNN regenerates training patterns better when the meta-prior has a high value. The four graphs show one dimension (y, here) of a training pattern (in orange) and the output regenerated by PV-RNN (in blue), obtained by bootstrapping the value of $\bm{A}_{1}$ with the one obtained during training, and computing predictions $\bm{X}_{1:T}$ using the generative model exclusively for the remaining timesteps. Black arrows point to the diverging steps in which the regenerated output diverges from the target pattern. }
\label{Fig5}
\end{figure}

During testing, the capability of the generative model to reproduce the training patterns was evaluated through target regeneration. For this purpose, target patterns were regenerated by providing the initial latent state $\bm{A}_{1}$  with the value obtained during training as we did in experiment 1.
The latent states $\bm{Z}_{2:T}$ and $\bm{X}_{1:T}$ were computed by the generative model. Figure~\ref{Fig5} illustrates how the regeneration is affected by different values of the meta-prior, and Table~\ref{tab3} shows the ADS for those values, i.e., in the continuous case, the timestep at which the mean square error between the target and the generated pattern exceeded a threshold ($0.01$) over 10 repetitions of each training sequence, as well as the mean activity of the variance $\bm{\sigma}^{(p)}$ for the whole training set. We obtain results in accordance with the ones of experiment 1: the PV-RNN model trained with the largest meta-prior value ($\wprime = 1.0$) exhibits deterministic dynamics, while the one trained with low values $\wprime$ approaches random process behavior.

\begin{table}[t!]
    \caption{ High meta-prior translates into deterministic dynamics, while low values produce random-process-like behavior. When $\wprime = 1.0$, the divergence starts from the $343th$ step driven by low stochasticity, and as $\wprime$ becomes smaller, the divergence starts earlier. When $\wprime = 0.01$, the divergence starts immediately after the onset and the variance is high.
   }
    \centering
    \begin{adjustbox}{max width=\textwidth}
    \begin{tabular}{c| c c c c c c}
    \cline{1-3}
    \toprule
           \multicolumn{1}{}  & &\multicolumn{6}{c}{$\wprime$} \\

           \multicolumn{1}{}  &  & $1.0$   & $0.5$  & $0.25$ & $0.15$ & $0.1$ & $0.01$\\
    \hline
            ADS& \textbf{343}  & 229 & 103 & 17 & 4 & 1\\
    \hline        
            Mean of Variance& \textbf{0.0007} & 0.0015 & 0.0039 & 0.005 & 0.021 & 0.1126\\ 
    \bottomrule
    \end{tabular}
    \end{adjustbox}
    \label{tab3}
\end{table}

\begin{table}[t!]
    \caption{ The best predictions are produced by the model trained with an intermediate value of the meta-prior ($\wprime = 0.25$). The table shows the MSE between the unseen test targets and the $1$-step to $5$-steps ahead generated predictions.}
    \centering
    \begin{adjustbox}{max width=\textwidth}
    \begin{tabular}{c| c c c c c c}
    \cline{1-3}
    \toprule
           \multicolumn{1}{}  & &\multicolumn{6}{c}{$\wprime$} \\

           \multicolumn{1}{}  &  & $1.0$   & $0.5$  & $0.25$ & $0.15$ & $0.1$ & $0.01$\\

    \hline
            1-step pred.& 0.0101  & 0.00726 & 0.00418 & 0.00376 & \textbf{0.00341} & 0.00834\\
    \hline
            2-steps pred.& 0.0171  & 0.0127 & \textbf{0.00907} & 0.00918 & 0.0116 & 0.0229\\
    \hline
            3-steps pred.& 0.0222  & 0.0183 & \textbf{0.0140} & 0.0152 & 0.0205 & 0.0378\\
    \hline
            4-steps pred.& 0.0279  & 0.0233 & \textbf{0.0189} & 0.0212 & 0.0301 & 0.0497\\
     \hline
            5-steps pred.& 0.0325  & 0.0274 & \textbf{0.0234} & 0.0270 & 0.0375 & 0.0578\\
    \bottomrule
    \end{tabular}
    \end{adjustbox}
    \label{tab4}
\end{table}

\begin{figure}[t!]
\hfill
\begin{center}
\includegraphics[width=5in]{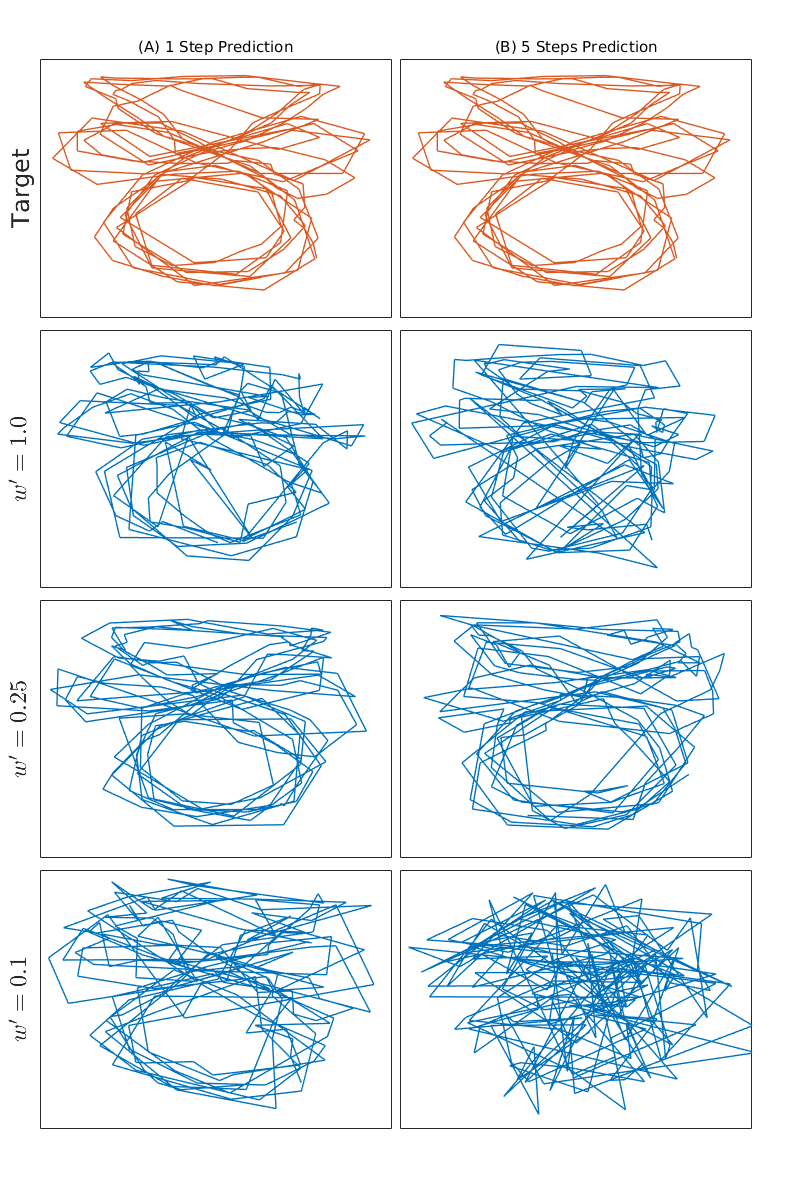}
\end{center}
\caption{ Only the $\wprime = 0.25$ case retains good qualitative behavior for both 1-step and 5-step ahead predictions. Target and prediction outputs of PV-RNNs with $\wprime$ set to $1.0$, $0.25$, and $0.1$ during error regression when predictions are made (A) one step ahead and (B) 5 steps ahead. Only the first 200 steps are shown to retain clarity.}
\label{Fig8}
\end{figure}

To test the generalization capabilities of the models, the prediction performance using error regression was evaluated. The test pattern of length $6400$ steps was given to each PV-RNN model to make predictions from $1$ to $5$ steps ahead. The size of the time-window was set to $50$ and $\bm{A}_{t-50:t-1}^{\textrm{test}}$ was optimized 30 times at every timestep. The time-window was continuously sliding one step forward at a time to generate the whole sequence $\bm{X}_{1:6400}$. The MSE between the test pattern and the generated output for all prediction steps is given in Table~\ref{tab4}. The PV-RNN trained with $\wprime$ set to $0.25$ outperforms other models in all cases except for the $1$-step ahead prediction where $\wprime = 0.1$ has a small lead. 2D visualizations of the test for 1-step and 5-step ahead predictions with $\wprime$ set to $1.0$, $0.25$, and $0.1$ are shown in Figure~\ref{Fig8}. As expected, predicting $5$-steps ahead is challenging for all models. However, in this case, the network with $\wprime$ set to $0.25$ performs best at preserving the structure of the target. When $\wprime$ is set to $0.1$, the network predicting 5 steps ahead generates a quite noisy pattern. The structure looks qualitatively wrong in some areas for both cases of prediction when $\wprime$ is set to $1.0$.

\begin{figure}[t!]
\hfill
\begin{center}
\includegraphics[width=5in]{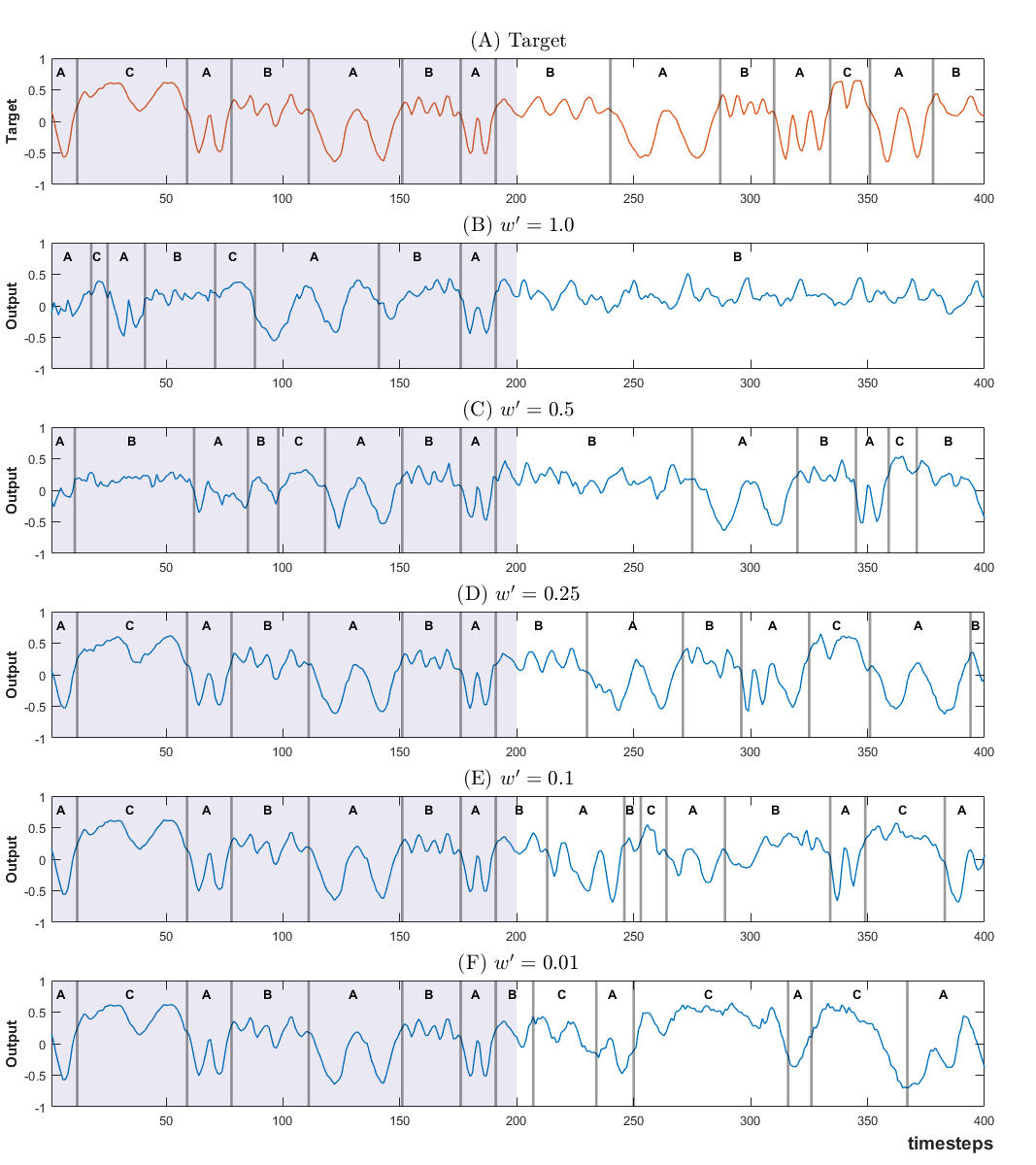}
\end{center}
\caption{$\wprime = 0.25$ produces the most faithful qualitative reproduction of sequence of primitives, both in terms of order and timing. It is the only one that reproduces the correct deterministic part of the target sequence $*, A, B, A, *, A, B$ (as well as producing possible stochastic steps (i.e., not producing $B$ on a stochastic step). In all these graphs, error regression is done once over the $[1,200]$ time-window, for 1000 optimization steps (shaded area). Then the generative model produces the remaining 200 timesteps, which are labelled to one of the three $A$, $B$, or $C$ primitives, based on their similarity to them. While here only the $y$ dimension is displayed, the labelling was done on the 2D signal. Gray bars display perceived transition primitive patterns (and actual ones for the target pattern).}
\label{Fig11}
\end{figure}

Previous performance measures focus on the model's ability to quantitatively reproduce or predict each timestep. To characterize the ability of the model to qualitatively reproduce and predict the correct patterns, we designed an experiment using error regression with a longer but fixed time-window. Contrary to previous experiments with error regression when the time-window would gradually grow to full size $m$ then slide over the whole test sequence, here we consider \emph{one} time-window starting at timestep 1 and ending at timestep 200 (included). In particular, we don't consider time-windows $[1, 1]$, $[1, 2]$, ..., $[1, 199]$. One thousand optimization steps of $\bm{A}_{1:200}^{test}$ are performed in this error regression time-window. Then the generative model is used to generate 200 additional steps, producing $\bm{X}_{201:400}$. This predicted 2D output is then analyzed and labelled by a human with the three primitive pattern types $A$, $B$, or $C$, and compared to the ground truth of the corresponding testing pattern. Figure~\ref{Fig11} shows one instance of the labelling of a test pattern, and illustrates the effects of different values of the meta-prior.

If a model produces the right primitive after the error regression window, it has 1-primitive prediction capability. If it produces the correct two primitive in the right order, it has 2-primitive prediction capability, and so on. Primitive reproduction is not rated on how long it lasts, or if it would coincide temporally with the test pattern. Table~\ref{tab5} shows the aggregated prediction performance of each PV-RNN model over the 32 patterns of the testing dataset. The prediction capability is best when $\wprime$ is set to $0.25$. This indicates that generalization in predicting at the primitive sequence level can be achieved to the highest degree by extracting the latent probabilistic structure in primitive sequences adequately when $\wprime$ balances well the two optimization terms of the lower bound. Previous works with MTRNN models have shown that higher layers  (middle and slow) can learn the transitions between primitive patterns while the lowest (fast) layer learns detailed information about primitive patterns  (\cite{yamashita2008emergence, ahmadi2017can}). Here, it is probable that PV-RNN was able to predict long-term primitive sequences by using the slow timescale dynamics developed in the higher layers of the network hierarchy.

\begin{table}[t!]
    \caption{Intermediate values of the meta-prior reproduce best the sequence of primitives. The table shows the percent accuracy for $1$, $2$, and $3$ primitive prediction for models trained with different values for $\wprime$.}
    \centering
    \begin{adjustbox}{max width=\textwidth}
    \begin{tabular}{c| c c c c c c}
    \cline{1-3}
    \toprule
           \multicolumn{1}{}  & &\multicolumn{6}{c}{$\wprime$} \\

           \multicolumn{1}{}  &  & $1.0$   & $0.5$  & $0.25$ & $0.15$ & $0.1$ & $0.01$\\
    \hline
            1-prim. pred. ($\%$)&   81.25 & 90.63 & \textbf{100} & 93.75 & 90.63 & 81.25\\
    \hline
            2-prim. pred. ($\%$)& 62.50  & 78.13 & \textbf{84.38} & 68.75 & 59.38 & 37.50\\
    \hline
            3-prim. pred. ($\%$)& 40.63  & 53.13 & \textbf{59.38} & 37.50 & 21.88 & 9.38\\
    \bottomrule
    \end{tabular}
    \end{adjustbox}
    \label{tab5}
\end{table}

\begin{figure}[t!]
\hfill
\begin{center}
\includegraphics[width=5in]{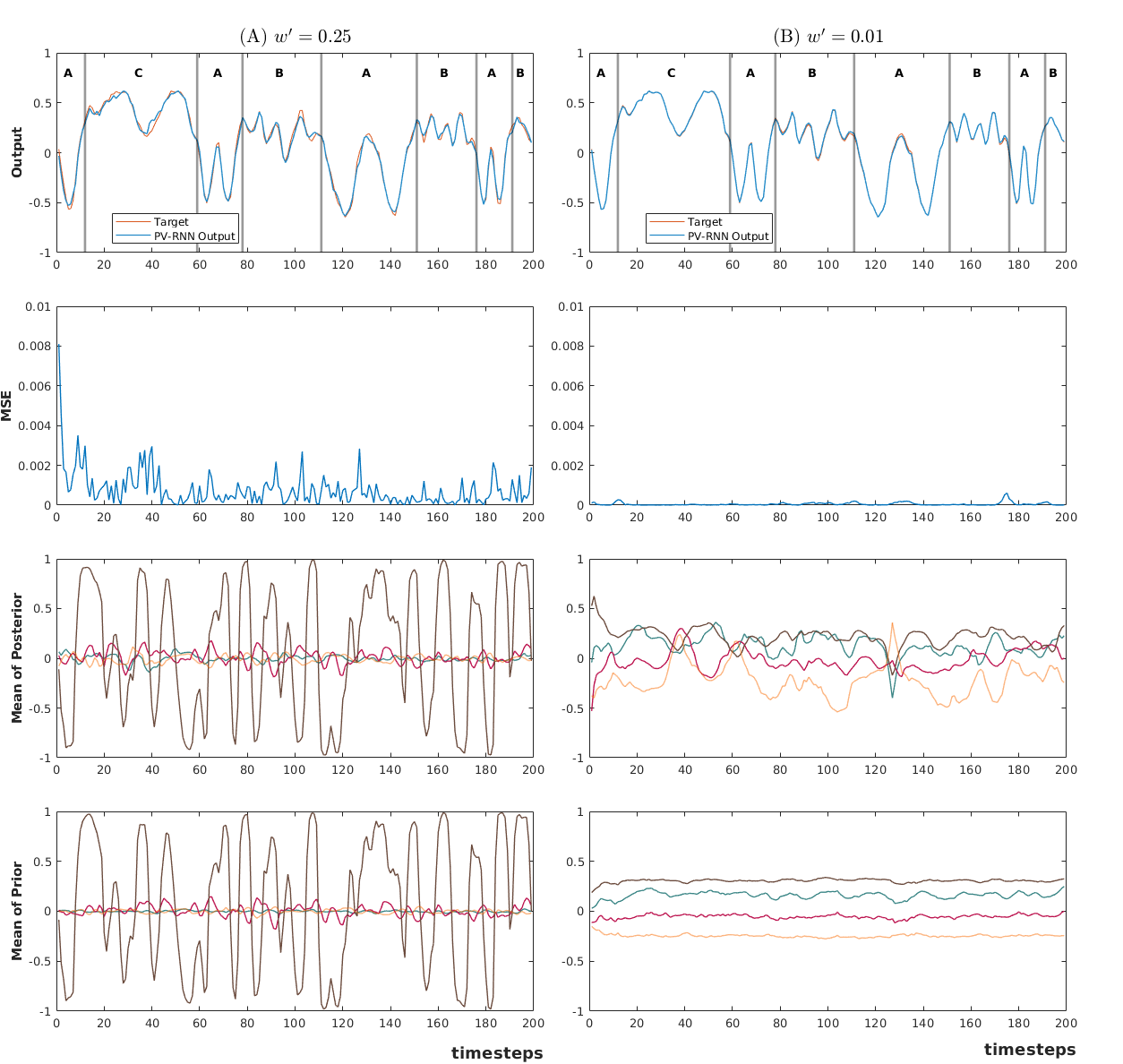}
\end{center}
\caption{ While the reconstruction is better for $\wprime = 0.01$ (lower MSE), the prior and posterior activity of the middle layer are significantly more different than for $\wprime = 0.25$, and the prior activity in particular seems to be constant. Here, we take a closer look at the activity of the $\wprime = 0.25$ and $\wprime = 0.01$ models presented in the previous figure (Figure \ref{Fig11}). Here, we consider only the part happening during the time-window $[1, 200]$. Only the $y$ dimension of the 2D patterns is illustrated in the first row of the graphs, however, the MSE (second row) is computed for both dimensions. The two bottom rows show the mean $\bm{\mu}^{(q)}_t$ (posterior) and $\bm{\mu}^{(p)}_t$ (prior) of the middle layer.}
\label{Fig14}
\end{figure}

One issue remains unclear in Figure~\ref{Fig11}: why did the predictability worsen even though the reconstruction becomes better when $\wprime$ goes from $0.25$ to $0.01$? To answer this, we compare the divergence between the posterior and the prior in the regression window of the $\wprime = 0.25$ and $\wprime = 0.01$ models. Figure~\ref{Fig14} shows the target patterns, the reconstructed outputs, the MSE between the target patterns and reconstructed outputs, the mean of posterior $\bm{\mu}^{(q)}_t$, and the mean of prior $\bm{\mu}^{(p)}_t$ for the middle layer, on the same test sequence. The reconstruction is more effective with $\wprime$ set to $0.01$, but the activities of the prior and posterior differ much more, as the activity of the mean of the middle layer indicates. Low values of the meta-prior lead to a low optimization pressure on the KL divergence term between the prior and posterior of the lower bound, and thus less pressure for the prior and posterior to coincide, and thus a poor learning for the prior (the activity of the mean of the prior for $\wprime = 0.01$ is poor), leading to poor prediction capabilities. Instead the optimization pressure concentrates on the reconstruction error term, leading to a lower MSE for the reconstruction. This analysis strongly suggests that a good balance between minimizing the reconstruction error and minimizing the KL divergence between the posterior and the prior by setting the meta-prior in an intermediate range is the best way to ensure the best performance of the network across a range of tasks. 

One question that may arise is how to find the optimal meta-prior at the beginning of the training. We don't have a good answer for that, as the optimal value depends on the task and the network topology. In the experiments of this paper, we conducted a simple grid search; this method does not guarantee finding the optimal value and is time-consuming. We may employ optimization techniques such as evolutionary algorithms, or consider the meta-prior as one of the training parameters of the network, and optimize it through backpropagation. Our preliminary experiments with learning the meta-prior did not show any satisfactory results and the network did not converge. We have left this issue as future work.


\section{Robot Experiment}
As explained in the previous section, PV-RNN was able to deal with probabilistic patterns during simulation experiments. We also conducted a robotics experiment involving synchronous imitation between two robots. This experiment allows us to do several things at once. One is to provide a more realistic test case, in higher dimensions (12), with complex perceptual noise source (e.g. hardware variations, motor noise, temperature, small synchronization discrepancies, and variations introduced when capturing training sequences from human movement). Another is to consider a context where actions need to be performed. The action space and sensory space are different in this experiment, leading us to adapt and apply the model in new ways. A final one is to compare PV-RNN with VRNN \parencite{chung2015recurrent} and thus to contrast the effectiveness of error regression versus an autoencoder-based approach. For this, synchronous imitation is considered to be an ideal task because it involves predicting both future perceptual sequences to compensate possible perceptual delay and recognizing others’ intention via posterior inference.



\subsection{Experimental Settings}
\begin{figure}[t!]
\hfill
\begin{center}
\includegraphics[width=5.8in]{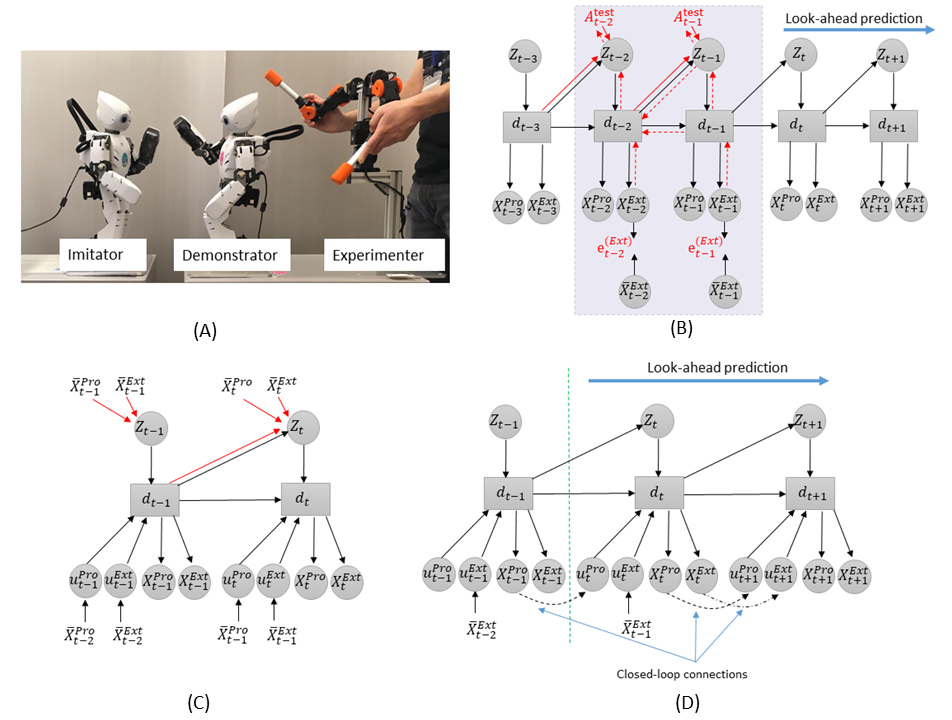}
\end{center}
\caption{Robotic experiment of synchronous imitation. One OP2 robot imitates the other OP2 tele-operated by the experimenter in (A), graph of computation and information flow using PV-RNN during testing (B), VRNN during training (C), and VRNN during testing (D). The red lines in (B) and (C) show the inference models of PV-RNN and VRNN, and dashed lines in (D) show how the closed-loop generation is computed. The models in (B), (C), and (D) are depicted with only one context layer for clarity. $\bm{u}_t^{Pro}$ and $\bm{u}_t^{Ext}$ represent the proprioception and exteroception input units at timestep $t$ that are fed by $\widebar{\bm{X}}_{t-1}^{Pro}$ and $\widebar{\bm{X}}_{t-1}^{Ext}$, respectively.}
\label{FigRobo}
\end{figure}

We used two identical OP2 humanoid robots placed face to face. One robot was the demonstrator, while the other was the imitator. To generate the training data, fifteen movement primitives of 200 timesteps, different from one another, were designed. A human then executed each primitive on both robots at the same time, in a mirror fashion, such that the movement of the left arm of one robot was executed on the right arm of the other (Figure~\ref{FigRobo} (A)). The imitator was then given training sequences composed of proprioception data---the joint angles of its own arms during movement, $\widebar{\bm{X}}_{t}^{Pro}$---and exteroception data---the XYZ coordinate of the hand tip of the demonstrator robot, from the perspective of the imitator robot, $\widebar{\bm{X}}_{t}^{Ext}$. 

The testing data was generated by a human, creating a movement sequence by repeating a movement primitive a few times (7 cycles on average) and then switching at random to another primitive, so that all fifteen primitives were used. The human strived to produce qualitatively the same movement, but not quantitatively: speed, amplitude and shape could differ. The resulting testing sequence is 4641 timesteps long.

During testing, the demonstrator would play the pre-recorded testing sequence, and the imitator robot would receive, at each timestep $t$, the corresponding exteroception data $\widebar{\bm{X}}_{t}^{Ext}$, but not, crucially, the $\widebar{\bm{X}}_{t}^{Pro}$ proprioception data. The imitator would use the PV-RNN model to make predictions about both the exteroception and the proprioception data. The proprioception predictions would be sent to the PID controller to be executed on the imitator robot, while the exteroception predictions would be compared with the actual observed one, $\widebar{\bm{X}}_{t}^{Ext}$, and the resulting error would be propagated through the network to perform error regression. Let's insist here that the errors propagated through the network are only relative to $\widebar{\bm{X}}_{t}^{Ext}$, as the target proprioception sequence is unavailable during testing. An important challenge in this setup is the switching, in the sequence, between different primitive patterns; the imitator must be able to recognize when they happen and update the internal state of the network, $\bm{A}^{\textrm{test}}$, appropriately (Figure~\ref{FigRobo} (B)). The same training and testing data was used on the same setup to train a VRNN model; Figure~\ref{FigRobo} (C) and (D) shows how VRNN was used in the training and testing, respectively.

To fairly compare the PV-RNN and VRNN, the same MTRNN type network layer structure was used for both networks. The internal dynamic of the MTRNN used in VRNN was computed as in Equation~\ref{eq:mtrnn} but for the lowest layer there is an added term $\bm{W}_{du} \bm{u}_{t}$ to feed the input $\bm{u}_{t}$ to the network. $\bm{u}_{t}$ is composed of $\bm{u}_{t}^{Pro}$ and $\bm{u}_{t}^{Ext}$. In that case, Equation~\ref{eq:mtrnn} becomes: 

\alignedequation{eq:mtrnn_VRNN}{
    \bm{h}_{t}^{1} &= (1 - \frac{1}{\tau_1})\bm{h}_{t-1}^{1} + \frac{1}{\tau_1}(\bm{W}^{1,1}_{dd} \bm{d}_{t-1}^1 + \bm{W}^{1,1}_{dz} \bm{Z}_{t}^1 + \bm{W}^{1,2}_{dd} \bm{d}_{t-1}^{2} +
    \bm{W}_{du} \bm{u}_{t})\\
    \bm{d}_{t}^1 &= \tanh(\bm{h}^1_{t})
}

In variational Bayes auto-encoder RNNs (\cite{chung2015recurrent, fraccaro2016sequential, goyal2017z, shabanian2017variational}), the previous timestep target $\widebar{\bm{X}}_{t-1}$ is provided as the current input to predict $\bm{X}_{t}$ in the output. As can be seen in Figure~\ref{FigRobo} (C), during training, $\bm{u}_{t}^{Pro} = \widebar{\bm{X}}_{t-1}^{Pro}$ and $\bm{u}_{t}^{Ext} = \widebar{\bm{X}}_{t-1}^{Ext}$. The look-ahead predictions for multiple steps were conducted by \emph{closed-loop generation} (\cite{tani1996model}) wherein both prediction of the proprioception and exteroception at a particular future timestep is obtained by feeding the prediction outputs of them at the previous timestep into the inputs at this step. Formally, during testing, for one step ahead prediction, we retain $\bm{u}_{t}^{Ext} = \widebar{\bm{X}}_{t-1}^{Ext}$ but the proprioception data comes from the prediction of the network itself, such that $\bm{u}_{t}^{Pro} = \bm{X}_{t-1}^{Pro}$. For two steps or more ahead predictions, we do not have access to exteroception targets. Therefore, both the proprioception and exteroception data come from the prediction of the network itself, such that, $\bm{u}_{t+1:T}^{Pro} = \bm{X}_{t:T-1}^{Pro}$ and $\bm{u}_{t+1:T}^{Ext} = \bm{X}_{t:T-1}^{Ext}$.

The approximate posterior of the VRNN is obtained as~(\cite{chung2015recurrent}):
\alignedequation{eq:posterior_distrib_VRNN}{
    q_{\phi}(\bm{Z}_{t}~|~\bm{d}_{t-1}, \widebar{\bm{X}}_{t}) = \mathcal{N}(\bm{Z}_{t}; \bm{\mu}_t^{(q)}, \bm{\sigma}_t^{(q)}) ~~~ \mathrm{where} ~~~ [\bm{\mu}_t^{(q)}, \log{\bm{\sigma}_t^{(q)}}]= f^{(q)}_{\phi}(\bm{d}_{t-1}, \widebar{\bm{X}}_{t})
}
where $f^{(q)}$, $\widebar{\bm{X}}$, and $\phi$ denote a one layer feed-forward neural network, the target, and the posterior variables, respectively. This means that in the VRNN (during training), the current timestep target (observation) is directly provided to the approximate posterior whereas in the PV-RNN not only the current timestep target but also future ones are indirectly provided to the approximate posterior through BPTT. The prior computation of VRNN is the same as Equation~\ref{eq:prior_distrib}.

We used the same number of $\bm{d}$ and $\bm{Z}$ units for the VRNN and PV-RNN models. Each model had three context layers consisting of $120$ $\bm{d}$ units and $12$ $\bm{Z}$ units for the fast context (FC) layer, $60$ $\bm{d}$ units and $6$ $\bm{Z}$ units for the middle context (MC) layer, and $30$ $\bm{d}$ units and $3$ $\bm{Z}$ units for the slow context (SC) layer. Time constants of FC, MC, and SC units were set to 2, 10, and 50, respectively. In each model case, the training ran for $50,000$ epochs. 

For testing, we performed prediction from 1 to 5 steps ahead. The motor controllers of the robot receive the last proprioception predictions as control targets: if we do one-step-ahead predictions, the motor controllers will receive $\bm{X}_{t}^{Pro}$ at time $t$. If we do five-step-ahead predictions, they will receive $\bm{X}_{t+4}^{Pro}$ at time $t$. This would allow to correct for perceptual and processing delay. For error regression in PV-RNN, the size of the time-window $m$ was set to $10$, and $100$ regression steps were performed at each timestep. Not providing VRNN with any proprioception information led to poor performance. To remedy this, we provided VRNN with 20 steps of proprioception information at the beginning of the test sequence. PV-RNN does not have access to any proprioception information.

Synchronized imitation with PV-RNN could not be done online because the error regression with 100 optimization steps costs about twice more computational time (205 ms) than real-time would allow. Therefore, in the case of using PV-RNN, the test was conducted on pre-recorded data sequence of exteroception. Although the VRNN case could be performed in real time, we also used the pre-recorded target sequences to ensure the most fair comparison.


\subsection{Experimental Results}

For PV-RNN and VRNN, we compared the performance for different values of the meta-prior $w$, namely $1.0$, $0.5$, $0.25$, $0.2$, and $0.1$. We present here the results for the best value of $w$ for each model, $0.5$ for PV-RNN and $0.25$ for VRNN.

The mean square error (MSE) between the target (exteroception and proprioception targets) and the predicted outputs (exteroception and proprioception outputs) with different look-ahead step length is shown for both cases of using PV-RNN and VRNN in Table~\ref{tab6}.
\begin{table}[t!]
    \caption{PV-RNN outperforms VRNN regardless of the number of prediction steps. The table shows the MSE between the target and look-ahead prediction with different timesteps ahead for PV-RNN and VRNN.}
    \centering  
    \begin{adjustbox}{max width=\textwidth}
    \begin{tabular}{c| c c c c c}    
    \cline{1-3} 
    \toprule
           \multicolumn{1}{}  & &\multicolumn{5}{c}{number of prediction steps} \\

           \multicolumn{1}{}  &  & 1 step   & 2 steps  &3 steps  & 4 steps & 5 steps \\             
    \hline
            PV-RNN &   0.00254 & 0.00259 & 0.00339 & 0.00395 & 0.00473\\
    \hline        
            VRNN & 0.00641  & 0.00702 & 0.00779 & 0.00852 & 0.00943\\
    \bottomrule
    \end{tabular}
    \end{adjustbox}
    \label{tab6}
\end{table}

In both PV-RNN and VRNN, the error increases as the model predict more steps ahead, as is expected. PV-RNN consistently outperforms VRNN, showing the effectiveness of error regression for prediction performance. 

We recorded two videos---1 step (\href{https://www.youtube.com/watch?v=gpGn8RVQtzM&t=10s}{Video1}) and 5 steps ahead predictions (\href{https://www.youtube.com/watch?v=38yU3Okb7Lk}{Video2})---of the movement patterns of three robots. One robot (middle) was the demonstrator robot and the other ones were the imitator robots: one controlled by the PV-RNN (left) and one by the VRNN (right). PV-RNN convincingly outperforms VRNN in the video of the 1-step ahead prediction. The PV-RNN robot synchronously imitates the target robot whereas the VRNN does not always perform a smooth, or correct, imitation. One step ahead prediction (50 ms) seems to be enough to overcome the perceptual delay in this setting, as delay is difficult to observe between demonstrator and imitator. In the 5-step ahead prediction, PV-RNN still shows better prediction performance than VRNN. However, it also fails to imitate the target robot for several movements, and exhibits brusk changes of speeds, accelerating and slowing down around the target pattern. Looking at the video of a successful imitation frame-by-frame, one can see that the imitator robot movements seem to be ahead of the target one.

\clearpage

\section{Discussion}

The current paper examines how uncertainty or probabilistic structure hidden in observed temporal patterns can be captured in an RNN through learning. To that end, it proposes a novel predictive-coding-inspired variational Bayes RNN. Our model possesses three main features that distinguish it from existing variational Bayes RNN. The first is the use of a weighting parameter, the meta-prior, between the two terms of the lower bound to control the optimization pressure. The second is propagating errors through backpropagation instead of propagating inputs during the forward computation. And the third is the error regression procedure during testing, performing online optimization of the internal state of the network during prediction. 

The idea of weighting the KL divergence term of the lower bound has been employed before, most notably in KL-annealing \parencite{bowman2015generating}. However in this paper, it is used for a different purpose. Through two experiments, the first one involving a finite state machine and the second one involving continuous temporal patterns composed of probabilistic transitions between a set of hand-generated movement, we showed that by changing the value of the meta-prior, we could achieve either a deterministic  or a random process behavior. The deterministic behavior reconstructed training sequences well and imitated the stochasticity of the data through deterministic chaos, but could not generalize to unseen testing sequences. The random process behavior could neither reconstruct training sequences nor generalize to unseen ones; its behavior was dominated by noise. The best value of the meta-prior could be found between those two extremes, where we showed that the network displayed both good reconstruction and generalization capability. It can be summarized that although probabilistic temporal patterns can be imitated by either deterministic chaos or stochastic process as suggested by the ergodic theory of chaos \parencite{crutchfield1992semantics}, the best representation can be developed by mixing the deterministic and stochastic dynamics via iterative learning of the proposed RNN model.

We employed the idea of propagating errors instead of propagating inputs to address two important issues in variational Bayes RNNs: how to provide the future dependency information to the latent states and how
to avoid ignoring the latent states during learning. We addressed both issues by introducing an adaptive vector $\bm{A}^{\widebar{\bm{X}}}$ in the inference model. $\bm{A}^{\widebar{\bm{X}}}$ is optimized through BPTT and captures the future dependencies of the external observation. This was verified in both simulated experiments, as providing the first timestep $\bm{A}_{1}^{\widebar{\bm{X}}}$ was sufficient to reconstruct the training sequences. Furthermore, because information from $\bm{A}^{\widebar{\bm{X}}}$ flows to $\bm{Z}$, and the information from $\bm{Z}$ flows through the deterministic states $\bm{d}$, which are ultimately responsible for the output sequences, the network is forced to construct good representations in the latent state $\bm{Z}$. This phenomenon was shown in the first experiment, when looking at the activity of mean $\bm{\mu}^{(p)}$ and variance $\bm{\sigma}^{(p)}$ of $\bm{Z}$.

Finally, we used an error regression procedure during testing, for making prediction about unseen sequences. In the second experiment, the PV-RNN based on a network architecture with multiple layers was evaluated for look-ahead prediction in primitive sequences. We found that relatively long sequences of primitive transitions were successfully predicted despite some discrepancies when $w$ was set to an adequate intermediate value. This, however, required a window of a sufficient length (200 steps), and a high number of iterations for error regression (1000 iterations). This suggests that a good balance between minimizing reconstruction error and divergence between the prior and posterior distributions can result in accurate prediction of future primitive sequences. 

Furthermore, the error regression procedure we employed bears a lot of similarities with the predictive coding principle. In particular, it shares the same processing cycle: making predictions, propagating prediction errors through the network hierarchy, and updating its internal state online to improve future predictions. Some important differences exist with predictive-coding implementations closer to the neurobiology of the brain \parencite{rao2000predictive}; in our model errors are propagated globally rather than locally. This is deliberate, to take advantage of the canonical tools of Variational Bayes RNNs. 

Other autoencoder-based variational Bayes RNNs infer the latent variable at each timestep through a recurrent mapping of the hidden state of the previous step, fed with inputs with the current timestep \parencite{fabius2014variational, bayer2014learning, chung2015recurrent}. In the robotic experiment for an imitation learning task, we showed that our model outperforms VRNN \parencite{chung2015recurrent}. This demonstrated the effectiveness of the error regression.   

The learning vector $\bm{A}^{\widebar{\bm{X}}}$ addressed two issues as explained above but introduces another. Indeed, as described in the inference model section, the dimension of $\bm{A}^{\widebar{\bm{X}}}$ increases linearly with the length and the number of training samples. It seems to preclude working with large datasets as a naive implementation might exceed any reasonable available memory. However, with each vector $\bm{A}^{\widebar{\bm{X}}}$ corresponding to a specific training pattern, it can be dynamically loaded and unloaded into memory whenever needed during training, resulting in a memory requirement equal to the one of the largest training batch. Furthermore, the trained values of the $\bm{A}^{\widebar{\bm{X}}}$ vectors are not needed for predicting unseen testing sequences and can be discarded entirely. If one wishes to perform reconstruction of the training patterns, only the first timestep $\bm{A}_{1}^{\widebar{\bm{X}}}$ is needed. Finally, although we did not do it in this paper, $\bm{A}^{\widebar{\bm{X}}}$ vectors may not be needed for every timestep and we could consider having $\bm{A}^{\widebar{\bm{X}}}$ vectors every 10 steps for instance. The implications are not necessarily trivial and this is a subject of ongoing study.

While the memory requirement of the model may not be a fundamental problem, perhaps a more serious issue lies in the computational requirement of the error regression process. Indeed, compared to most models that only need forward computations for evaluation, our model still needs to backpropagate and perform optimization. This can severely hamper its ability to be deployed on a variety of platforms. It can also be an issue for real-time robotics, as was the case in our robotic experiment. We are currently investigating ways to reduce the computational burden of error regression.  

An intriguing consideration is that the current results showing that the generalization capability of PV-RNN depends on the setting of the meta-prior $w$ bears parallels to observational data about autism spectrum disorders (ASD) and may suggest possible accounts of its underlying mechanisms. ASD is a wide-ranging pathology including deficits in communication, abnormal social interactions, and restrictive and/or repetitive interests and behaviors \parencite{dicicco2006qualitative}. Recently, there has been an emerging view suggesting that deficits in low-level sensory processing may cascade into higher-order cognitive competency, such as in language and communication \parencite{stevenson2014multisensory, lawson2014aberrant, robertson2017sensory}. \textcite{van2014precise} have suggested that ASD might be caused by overly strong top-down prior potentiation to minimize prediction errors (thus increasing precision) in perception, which can enhance capacities for rote learning while resulting in the loss of the capacity to generalize what is learned, a common ASD symptom.

This account by \textcite{van2014precise} corresponds to some extent to the situation of PV-RNN when learning with $w$ set to a larger value. PV-RNN is able to exactly reconstruct complex training sequences by embedding them in deterministic dynamics, while being unable to generalize to unseen sequences. With a larger value of $w$, the optimization pressure on the KL divergence term, and therefore on the prior to be similar to the posterior, is stronger. This pushes the network to reduce uncertainty in the network (low $\bm{\sigma}$), increasing the precision of the predictions, resulting in a stronger top-down prior. Such a phenomenon could explain how ASD patients in social contexts might frequently suffer from over-amplified error in predicting behaviors of others; this results from  overestimated precision in prediction due to overfitting in learning. For this reason, such patients may tend to indulge in their own repetitive behaviors that generate a tolerable amount of error. 
Then, mechanisms akin to the inability of adapting $w$ adequately may result in the pathology. \textcite{lawson2014aberrant} proposes that maladaptation of precision itself in hierarchical message passing in the brains may contribute to many features of autistic perception in a study using a hierarchical Bayesian model built on the predictive coding framework. If a neural mechanism indeed exists that corresponds to adapting $w$, study about the precise feedback mechanisms to regulate $w$ within an optimal range could be an important research question for ASD. It could also shed light on how the brain handles various cognitive tasks using statistical inference. 

Various robotics applications would be interesting for future study. One of the main features of the proposed model is that it can learn the latent probabilistic structure of data; this should translate into high performance at learning skilled behaviors through supervised teaching such as manual manipulation of objects. Indeed, a crucial component in generating skilled behavior is that precision in movement control must change depending on the situation during task execution. For example, when grasping an object, the precision during reaching can be low, but it must become much higher at the moment of contact with the object, to establish a good grasp. It is highly likely that PV-RNN can learn to extract such statistical structure by inferring the necessary precision in generating movement trajectories from the set of training trajectories.

Another interesting direction for future study would be the introduction of goal-directed planning mechanisms into the current model. Arie and colleagues (\cite{arie2009creating}) showed that the deterministic MTRNN model can generate goal-directed plans. It does that by starting from a desired goal state for a future timestep, and backpropagating the error to infer the necessary context state of the current timestep to achieve this goal. From this inferred initial context state, a proprioception sequence (joint angles of the robot) can be predicted, and used to directly specify what actions the robot should undertake. Furthermore, Butz and  colleagues (\cite{butz2019learning}) recently proposed REPRISE, a REtrospective and PRospective Inference SchEme, which can infer both retrospectively for past contextual event states and prospectively for future optimal motor command sequences satisfying some given goal states. REPRISE is built on their previous work which included the prospective phase using an RNN \parencite{otte2017inherently}. In the new proposed model, an RNN is augmented with contextual neurons in order to encode continuous sensorimotor dynamics into sequences of discrete events. The contextual neurons are adapted during the retrospective phase in order to minimize the loss between predicted and actual sensory information. Then, the motor commands in the future timesteps are adapted via BPTT during the prospective phase in order to minimize the discrepancies between predicted future states and desired goal states. Such a mechanism could also be realized in PV-RNN by inferring the optimal adaptive vector $\bm{A}$ sequence accounting both for past sensory experience and for the specified future goal states. Future study should examine how the extended PV-RNN can perform goal-directed planning tasks including online replanning ones, compared to existing models such as REPRISE \parencite{butz2019learning}. 

The robotic experiment presented in this paper was limited in many ways. We are now working on a two-robot setup where there is not one demonstrator and one imitator, but two imitators imitating each other. In the context of active inference \parencite{friston2009reinforcement, friston2010action, pezzulo2015active, baltieri2017active}, an agent interacting with an environment has two choices when its predictions don't agree with its observations: either modify its internal state to produce predictions that better align with observations, or perform an adequate intervention in the environment to make observations better correspond to the predictions. In other words, when the world does not fit our expectations, we can change our expectations, or change the world by acting adequately on it. In a context where a demonstrator performs for an imitator, the imitator has no choice but to change its internal state when prediction errors occur. But in a situation with two imitators, robot A may learn how robot B responds to its own actions, especially when robot A's actions generate prediction errors for robot B. If robot B is only updating its internal state, robot A might end up continuously performing interventions on robot B's behavior rather than changing its own internal state. The most interesting case should happen when both robots are able to learn to predict the consequences of their own actions on the other robots \emph{and} perform interventions to influence one another's actions. The circular causality developed between those two may lead to ambiguity in determining which one drives the other, which one demonstrates, and which one imitates, with possibly continuous switching between those roles. The early results we obtained on such as setup are encouraging. Studies examining such aspects could greatly contribute to understanding of the underlying mechanisms of social cognition and how to  engineer the autonomous development of collaborative actions among multiple agents.


\clearpage

\section{Conclusion}

We proposed a predictive-coding inspired variational Bayes RNN to capture the stochasticity of time-series data. To that end, the cost function of the network is composed of two terms: the expected prediction error and the KL-divergence (measuring how similar two distributions are) between the posterior and prior distributions. The relative importance of those two terms is weighted by a parameter $w$, the meta-prior.

First, in a simple task, we demonstrated that increasing the value of the metaprior $w$, and therefore, the optimization pressure on  the KL divergence term in the cost function, led to the model behaving increasingly deterministically, leading to development of deterministic chaos, with low generalization capabilities. Conversely, lowering the value of $w$ led to a network behaving increasingly stochastically. Stochastic models are better at generalization, but, at the extreme, turn into random generators that disregard the structure of the input data.

The best behavior is found when the value of $w$ is between those two extremes:  the network was able to achieve the best generalization capability with an intermediate value of the meta-prior. We confirmed this observation on a more complex tasks using both hand-drawn patterns and robotic motions, and on a more complex model using a higher number of context layers.

Our approach provides interesting solutions to two issues variational Bayes RNNs typically have: latent variables are ignored during training, and they do not have access to information about the future dependencies of the data. Our network solves those two issues by avoiding to feed inputs to the network during the forward computation, preferring to propagate prediction errors during BPTT to dedicated latent variables.

These variables are leveraged when predicting unseen testing sequences: we use an error regression procedure during evaluation, which performs online optimization of the internal state of the network based on observed prediction errors. Therefore, the predictions are  constantly re-evaluated as new external data becomes observable. We have shown that our model outperforms the VRNN model \parencite{chung2015recurrent} on a robotic imitation task.

\section*{Acknowledgement}
We would like to give our special thanks to people who helped us with the current study. 
First and foremost, we are particularly grateful for great assistant and insightful advice given by Fabien Benureau for improving the content and language of the paper. We sincerely express our appreciation to Tom Burns, Nadine Wirkuttis, Wataru Ohata, Takazumi Matsumoto, and Siqing Hou for their great help improving this work as well. 
\clearpage

\section*{Appendix}

\setcounter{equation}{0}
\setcounter{subsection}{0}
\renewcommand{\thesubsection}{\Alph{subsection}}

\renewcommand\thefigure{A\arabic{figure}}    
\setcounter{figure}{0}

\setcounter{table}{0}
\renewcommand{\thetable}{A\arabic{table}}

\subsection{Posterior Computation}\label{Posterior}

\begin{equation} \label{eq:appA_1}
    q_{\phi}(\bm{Z}_{t}~|~\bm{d}_{t-1}, \bm{e}_{t:T}) = \mathcal{N}(\bm{Z}_{t}; \bm{\mu}_t^{(q)}, \bm{\sigma}_t^{(q)}) ~~~ \mathrm{where} ~~~ [\bm{\mu}_t^{(q)}, \log{\bm{\sigma}_t^{(q)}}]= f^{(q)}(\bm{d}_{t-1}, \bm{A}_{t}^{\widebar{\bm{X}}})
\end{equation}

The equation for $\bm{\mu}$ and $\log \bm{\sigma}$ of posterior can be written as
\begin{equation} \label{eq:appA_2}
    \left\{
    \begin{array}{ll}
      \bm{\mu}_{t}^{k} = \tanh{(\bm{W}_{\mu d}^{kk} \tilde{\bm{d}}_{t-1}^{k} + \bm{A}_{\mu, t}^{\widebar{\bm{X}},k})}\\
      \log{\bm{\sigma}_{t}^{k} = \bm{W}_{\sigma d}^{kk} \tilde{\bm{d}}_{t-1}^{k} + \bm{A}_{\sigma, t}^{\widebar{\bm{X}},k}}
    \end{array}
  \right.
\end{equation}
where $\bm{\mu}_{t}^{k}$ is the vector of the mean values of the $k_{th}$ context layer at time $t$, $\bm{W}_{\mu d}^{kk}$ is the matrix of the connectivity weights from the $\bm{d}$ units in the $k_{th}$ context layer to $\bm{\mu}$ units in the same context layer and $\bm{W}_{\sigma d}^{kk}$ is the matrix of the connectivity weights from the $\bm{d}$ units in the $k_{th}$ context layer to the $\bm{\sigma}$ units in the same context layer. The notation \enquote{$(q)$} was omitted from the equation for the sake of simplicity. $\bm{A}_{\mu, t}^{\widebar{\bm{X}},k}$ and $\bm{A}_{\sigma, t}^{\widebar{\bm{X}},k}$ are obtained as follows
\begin{equation} \label{eq:appA_3}
    \left\{
    \begin{array}{ll}
      \bm{A}_{\mu, t}^{\widebar{\bm{X}},k} = \bm{A}_{\mu, t}^{\widebar{\bm{X}},k} + \alpha \frac{\partial L}{\partial \bm{A}_{\mu, t}^{\widebar{\bm{X}},k}}\\
      \bm{A}_{\sigma, t}^{\widebar{\bm{X}},k} = \bm{A}_{\sigma, t}^{\widebar{\bm{X}},k} + \alpha \frac{\partial L}{\partial \bm{A}_{\sigma, t}^{\widebar{\bm{X}},k}}
    \end{array}
  \right.
\end{equation} 
where $\alpha$ is the learning rate. Based on Eq.~\ref{eq:appA_2}, we can rewrite Eq.~\ref{eq:appA_3} to have the derivatives with respect to mean and standard deviation values as
\begin{equation} \label{eq:appA_4}
    \left\{
    \begin{array}{ll}
      \bm{A}_{\mu, t}^{\widebar{\bm{X}},k} = \bm{A}_{\mu, t}^{\widebar{\bm{X}},k} + \alpha~(1 - \tanh^2{(\bm{W}_{\mu d}^{kk} \tilde{\bm{d}}_{t-1}^{k} + \bm{A}_{\mu, t}^{\widebar{\bm{X}},k})})~(\frac{\partial L}{\partial \bm{\mu}_{t}^{k}})\\
      \bm{A}_{\sigma, t}^{\widebar{\bm{X}},k} = \bm{A}_{\sigma, t}^{\widebar{\bm{X}},k} + \alpha~\frac{\partial L}{\partial \log\bm{\sigma}_{t}^{k}}
    \end{array}
  \right.
\end{equation}  

\setcounter{equation}{0}
\subsection{KL Divergence}\label{KL}

We let each posterior and prior distributions be a Gaussian with a diagonal covariance matrix, so:

\small
\begin{equation} \label{eq:appB_1}
    KL[q_\phi(\bm{Z}_{t})~||~P_{\theta_Z}(\bm{Z}_{t})] = E_{q_\phi}[\log{q_\phi(\bm{Z}_{t})}] - E_{q_\phi}[\log{P_{\theta_Z}(\bm{Z}_{t})}]
\end{equation}
\normalsize

\small
\begin{equation} \label{eq:appB_2}
    q_\phi(\bm{Z}_{t}) = \frac{1}{\sqrt{2\pi(\bm{\sigma}_t^q)^2}} e^{\frac{-(\bm{Z}_{t} - \bm{\mu}_t^q)^2}{2(\bm{\sigma}_t^q)^2}}
\end{equation}
\normalsize

\small
\begin{equation} \label{eq:appB_3}
    P_{\theta_Z}(\bm{Z}_{t}) = \frac{1}{\sqrt{2\pi(\bm{\sigma}_t^p)^2}} e^{\frac{-(\bm{Z}_{t} - \bm{\mu}_t^p)^2}{2(\bm{\sigma}_t^p)^2}}
\end{equation}
\normalsize
For simplicity, we removed parenthesis from $p$ and $q$. Based on Equations~\ref{eq:appB_1}, \ref{eq:appB_2}, and \ref{eq:appB_3}:

\small
\begin{equation} \label{eq:appB_4}
    E_{q_\phi}[\log{q_\phi(\bm{Z}_{t})}] = E_{q_\phi}[-\frac{1}{2}\log{2\pi}-\frac{1}{2}\log{(\bm{\sigma}_t^q)^2}+\frac{-(\bm{Z}_{t} - \bm{\mu}_t^q)^2}{2(\bm{\sigma}_t^q)^2}]
\end{equation}
\normalsize

\small
\begin{equation} \label{eq:appB_5}
\begin{split}
    E_{q_\phi}[\log{P_{\theta_Z}(\bm{Z}_{t})}] &= -\frac{1}{2} E_{q_\phi} [\log{2\pi}+\log{(\bm{\sigma}_t^p)^2}+\frac{(\bm{Z}_{t} - \bm{\mu}_t^p)^2}{(\bm{\sigma}_t^p)^2}] \\ 
    &= -\frac{1}{2} E_{q_\phi}[\log{2\pi}+\log{(\bm{\sigma}_t^p)^2}+\frac{(\bm{Z}_{t})^2 + 2\bm{Z}_{t}\bm{\mu}_t^p - (\bm{\mu}_t^p)^2}{(\bm{\sigma}_t^p)^2}]
\end{split}
\end{equation}
\normalsize

Variance and $E[\bm{Z}_{t}^2]$ can be written as:
\small
\begin{equation} \label{eq:appB_6}
    \bm{\sigma}_t^2 = E[(\bm{Z}_{t} - \bm{\mu}_t)^2],~~~~~~E[\bm{Z}_{t}^2] = \bm{\mu}_t^2 + \bm{\sigma}_t^2
 \end{equation}
\normalsize
So, Equations~\ref{eq:appB_4} and~\ref{eq:appB_5} can be rewritten as:
\small
\begin{equation} \label{eq:appB_7}
\begin{split}
    E_{q_\phi}[\log{q_\phi(\bm{Z}_{t})}] &= -\frac{1}{2}\left(\log{2\pi}+\log{(\bm{\sigma}_t^q)^2}+\frac{(\bm{\sigma}_t^q)^2}{(\bm{\sigma}_t^q)^2}\right) \\
    &= -\frac{1}{2}\left(\log{2\pi}+\log{(\bm{\sigma}_t^q)^2}+1\right)
\end{split}
\end{equation}
\normalsize
\small
\begin{equation} \label{eq:appB_8}
    E_{q_\phi}[\log{P_{\theta_Z}(\bm{Z}_{t})}] = -\frac{1}{2}\left(\log{2\pi} + \log{(\bm{\sigma}_t^p)^2}+\frac{(\bm{\mu}_t^q)^2 + (\bm{\sigma}_t^q)^2 - 2\bm{\mu}_t^q\bm{\mu}_t^p + (\bm{\mu}_t^p)^2}{(\bm{\sigma}_t^p)^2}\right)
\end{equation}
\normalsize
Now, Equation~\ref{eq:appB_1} can be rewritten as:
\small
\begin{equation} \label{eq:appB_9}
\begin{split}
    KL[q_\phi(\bm{Z}_{t})~||~P_{\theta_Z}(\bm{Z}_{t})] &= -\frac{1}{2}\left(\log{(\bm{\sigma}_t^q)^2}+ 1 -\log{(\bm{\sigma}_t^p)^2} - \frac{(\bm{\mu}_t^q)^2 +(\bm{\sigma}_t^q)^2 - 2\bm{\mu}_t^q\bm{\mu}_t^p + (\bm{\mu}_t^p)^2}{(\bm{\sigma}_t^p)^2} \right)\\
    &= \log{\frac{\bm{\sigma}_t^{(p)}}{\bm{\sigma}_t^{(q)}}}+\frac{(\bm{\mu}_t^{(p)} - \bm{\mu}_t^{(q)})^2 + (\bm{\sigma}_t^{(q)})^2}{2(\bm{\sigma}_t^{(p)})^2} -\frac{1}{2}
\end{split}
\end{equation}
\normalsize


\setcounter{section}{-1}
\subsection{Lyapunov Exponent Computation}\label{lya}

For the sake of simplicity, let us consider a PV-RNN consisting of 2 $\bm{d}$ units and $1$ $\bm{Z}$ unit. We need to first compute Jacobian matrices at each timestep as
\[
J_t =
\begin{bmatrix}
    \frac{\partial{\bm{Z}_{t+1,1}}}{\partial{\bm{Z}_{t,1}}}   & \frac{\partial{\bm{Z}_{t+1,1}}}{\partial{\bm{d}_{t,1}}} & \frac{\partial{\bm{Z}_{t+1,1}}}{\partial{\bm{d}_{t,2}}}\\
    \frac{\partial{\bm{d}_{t+1,1}}}{\partial{\bm{Z}_{t,1}}}   & \frac{\partial{\bm{d}_{t+1,1}}}{\partial{\bm{d}_{t,1}}} & \frac{\partial{\bm{d}_{t+1,1}}}{\partial{\bm{d}_{t,2}}} \\
    \frac{\partial{\bm{d}_{t+1,2}}}{\partial{\bm{Z}_{t,1}}}   & \frac{\partial{\bm{d}_{t+1,2}}}{\partial{\bm{d}_{t,1}}} & \frac{\partial{\bm{d}_{t+1,2}}}{\partial{\bm{d}_{t,2}}}
\end{bmatrix}
\]
It can be noted that $\bm{X}$ does not exist in the Jacobian matrices because in the generative model, $\bm{X}_{1:T}$ are not given to the context layers. We can now resort the approximation of the image ellipsoid $J_T J_{T-1} ... J_{1} U$ of the unit sphere by a computational algorithm. More details can be seen in (\cite{alligood1996chaos}). Here, the computation of the first-largest Lyapunov exponent is only given.  Let us start with an orthonormal basis $r$ = $[1.0~0.0~0.0]^T$, and use the Gram-Schmidt orthogonalization procedure, so we have

\begin{algorithm}
  \caption{Lyapunov Exponent Computation}\label{euclid}
  \begin{algorithmic}[1]
  \State  $LE = 0.0$
      \For{\texttt{<t = 1 to T>}}
        \State $y_t = J_t~r$
        \State  $r = \frac{y_t}{\lVert \mathbf{y_t} \rVert}  $
        \State  $LE~\pluseq~\log{\lVert \mathbf{y_t} \rVert}  $
      \EndFor
      \State  $LE =  \frac{LE}{T}$
  \end{algorithmic}
\end{algorithm}
\noindent where $\lVert . \rVert$ and $LE$ denote Euclidean length, and first largest Lyapunov exponent, respectively. $T$ was $50, 000$ in our experiments.

\clearpage

\begin{figure}[ht!]
\hfill
\begin{center}
\includegraphics[width=5in]{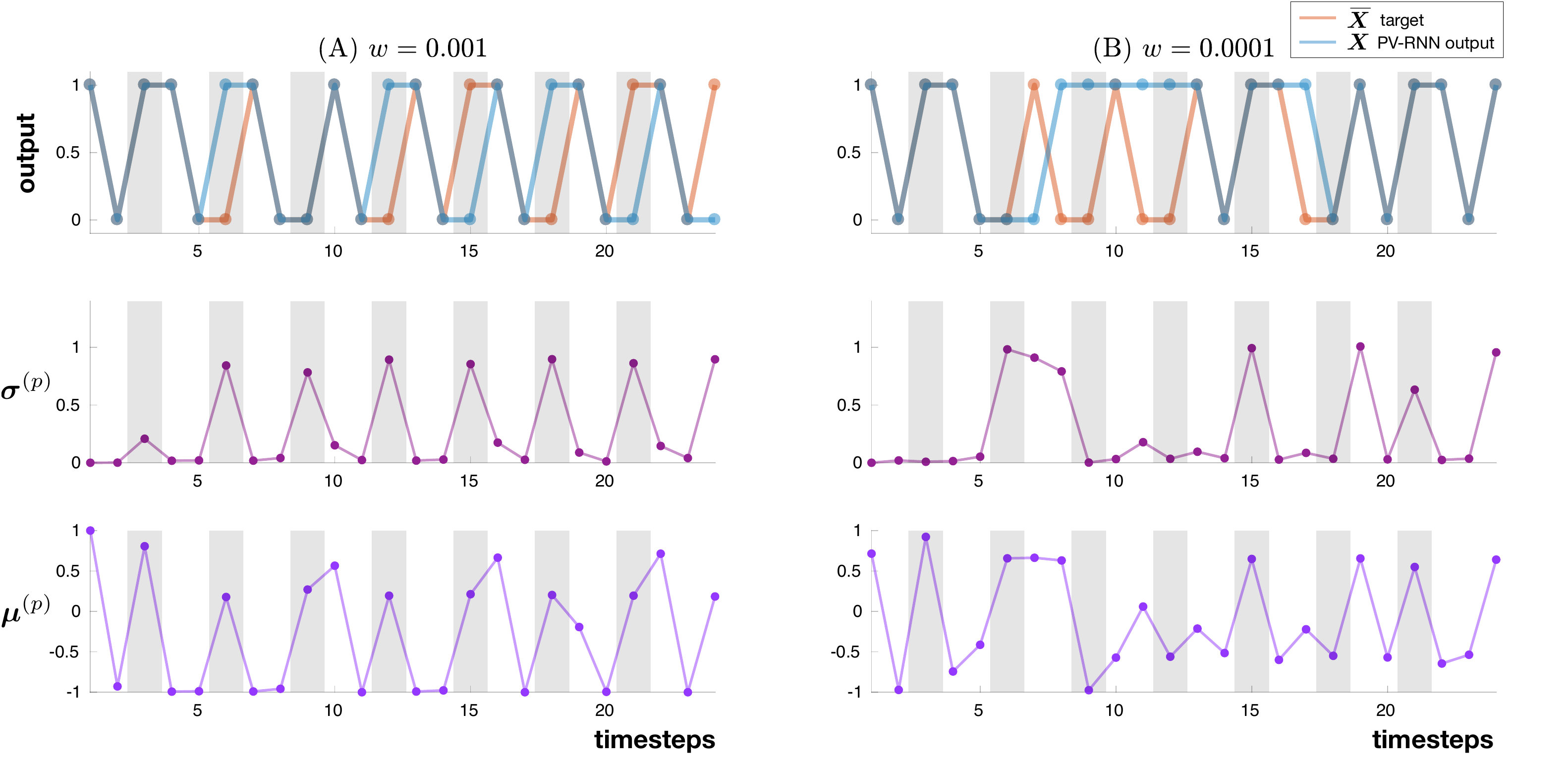}
\end{center}
\caption{ The target and the regenerated outputs, the mean $\bm{\mu}^{(p)}$, and the standard deviation $\bm{\sigma}^{(p)}$ of two PV-RNNs trained with meta-prior $w$ set to $0.01 \times 10^{-1}$ (A) and $0.001 \times 10^{-1}$ (B). The gray bars show the timesteps corresponding to uncertain states.}
\label{AppFig2}
\end{figure}

\begin{figure}[ht!]
\hfill
\begin{center}
\includegraphics[width=5in]{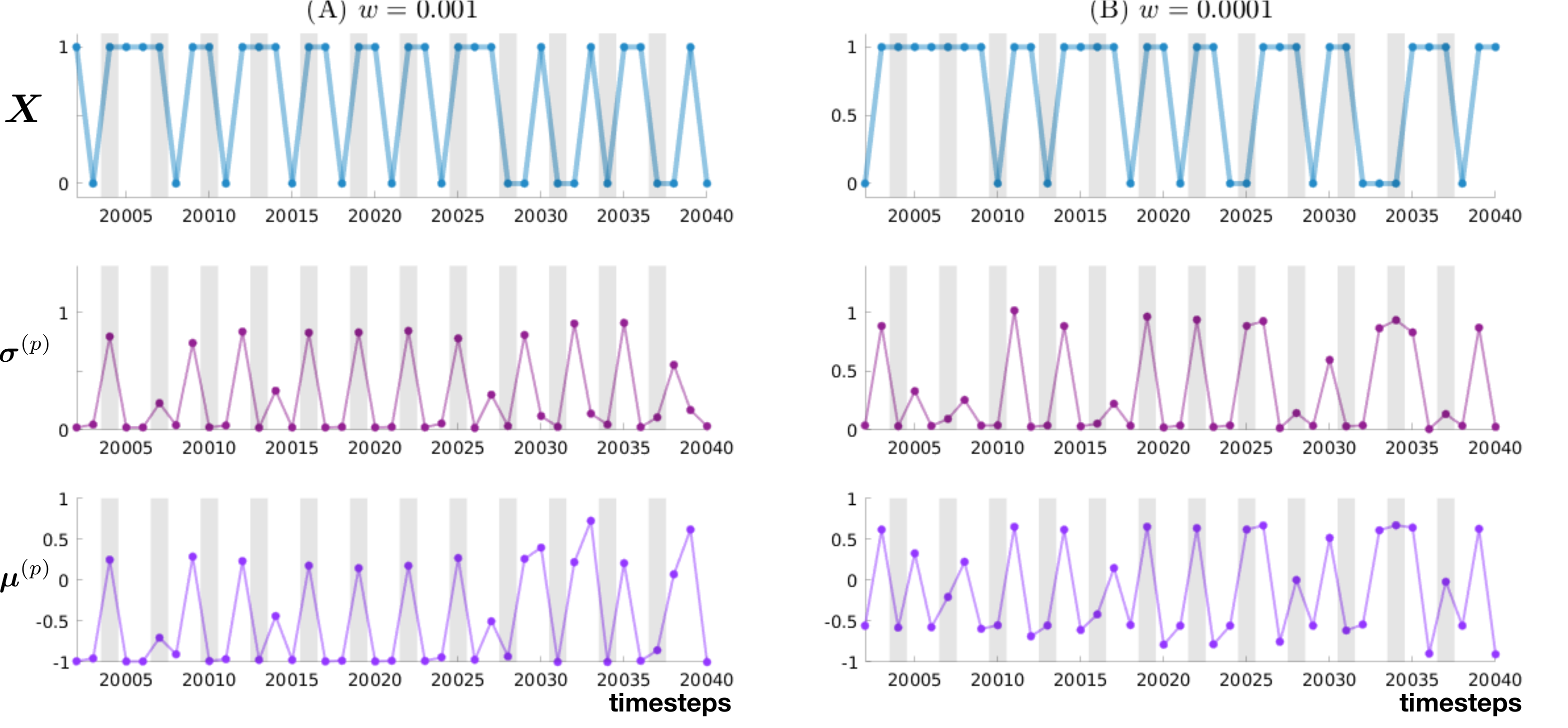}
\end{center}
\caption{ The generated output, the mean $\bm{\mu}^{(p)}$, and the standard deviation $\bm{\sigma}^{(p)}$ from timesteps $20,002$ to $20,040$ of two PV-RNNs trained with the meta-prior $w$ set to $0.01 \times 10^{-1}$ (A) and $0.001 \times 10^{-1}$ (B). The gray bars show the timesteps corresponding to uncertain states.}
\label{AppFig3}
\end{figure}

\begin{figure}[ht!]
\hfill
\begin{center}
\includegraphics[width=5in]{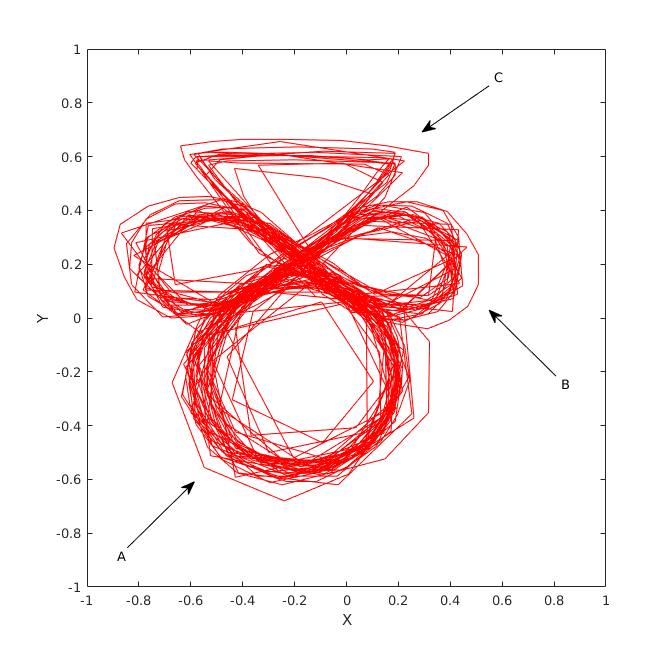}
\end{center}
\caption{ The A, B, and C patterns generated based on PFSM shown in Figure \ref{AppFig1}(B). The whole pattern is 1000 timesteps including 20 of pattern A, 13 of pattern B, and 7 of pattern C: each pattern has 2 cycles. The patterns are not identical and contain fluctuations in amplitude, velocity, and shape.}
\label{AppFig4}
\end{figure}

\begin{figure}[ht!]
\hfill
\begin{center}
\includegraphics[width=6in]{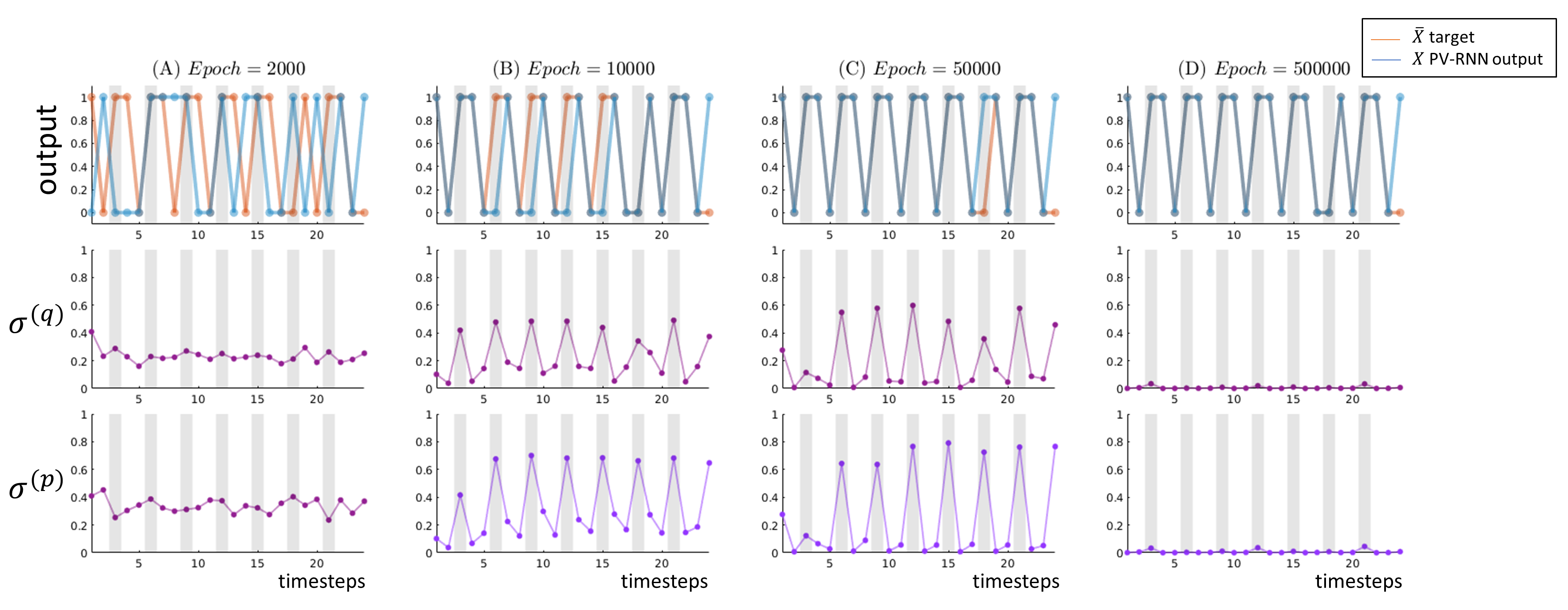}
\end{center}
\caption{ At the early stage of the training, $\bm{\sigma}^{(q)}$ and $\bm{\sigma}^{(p)}$ are large for both the probabilistic states (gray bars) and the deterministic states: the transition rules defined in the PFSM are mostly broken. After $10000$ epochs, $\bm{\sigma}^{(q)}$ and $\bm{\sigma}^{(p)}$ are significantly higher for the probabilistic states than the deterministic ones. Therefore, the reconstruction of the training patterns are successfully done on the deterministic states. At the final stage of the training, $\bm{\sigma}^{(q)}$ and $\bm{\sigma}^{(p)}$ are near zero and the target sequence is completely regenerated. These results show that the network reconstructs the deterministic states of the training patterns first. }
\label{AppFig5}
\end{figure}

\clearpage
\subsection{Experiment 1 with a Simple RNN}
\label{appendix:exp1simpleRNN}
We conducted Experiment 1 again using the same datasets and network parameter settings with the exception of the time constant. We set the time constant in Equation \ref{eq:mtrnn} to 1.0, which removes the leaky integrator and the $\bm{d}$ unit becomes a simple RNN. ADS and KL divergence of the test phase for different values of $w$ are shown in Table \ref{Apptab1}. The results are in line with the ones shown in Section \ref{Exp1} where the MTRNN with time constant 2.0 was used.   

\begin{table}[ht]
    \caption{The Average Diverging Step (ADS) points to better reconstruction performance when $w$ is high. However, taking into account the KL divergence between the probabilistic distribution of the generated pattern $P(\bm{X}_{t:t+11})$ and the one of the training data $P(\widebar{\bm{X}}_{t:t+11})$ paints another picture: the network best captures the probabilistic structure of the data for an average value of $w$.}
    \centering  
    \begin{adjustbox}{max width=\textwidth}
    \begin{tabular}{c| c c c c c c c}    
    \cline{1-3} 
    \toprule
           \multicolumn{1}{}  & &\multicolumn{7}{c}{Meta-Prior $w$} \\ 
        
           \multicolumn{1}{}  &  & $0.1$   & $0.05$  & $0.025$ & $0.015$ & $0.01$ & $0.001$ & $0.0001$\\  
           
    \hline
            Average Diverging Step (ADS) & \textbf{23}  & 19 & 15 & 12 & 11 & 8 & 7\\
    \hline        
            KL div. of Test Phase & 6.8589  & 1.6515 & \textbf{0.1106} & 0.1306 & 0.359 & 0.8758 & 2.8845\\
    \bottomrule
    \end{tabular}
    \end{adjustbox}
    \label{Apptab1}
\end{table}

\subsection{Experiment 2 Dataset}
\label{appendix:exp2data}

The dataset for experiment 2 data sequences was generated in three stages as follows.

First, three different 2D movement primitive patterns and a PFSM of defining the transition probability among those patterns were prepared. Then, a human subject was asked to draw patterns in 2D using a tablet device by sequentially concatenating the primitive patterns by following the transition probability defined in the PFSM of Figure~\ref{AppFig1}(B).
Each drawing of hand drawing primitive patterns necessarily contains fluctuations in amplitude, velocity, and shape. Primitive patterns \enquote{A, B, C} were circles, rotated figure-eight, and triangles similar to those depicted in the first row of Figure~\ref{Fig8}. Each of them is a cyclic pattern with periodicity 2.
The PFSM adopted in this experiment is shown in  Figure~\ref{AppFig1}(B).
The total number of \enquote{A, B, C} primitive patterns generated was 160 (4458 timesteps). The branching after $s_4$ either by generating a primitive $B$ or $C$ was randomly chosen by the human. We measured the conditional probabilities in the data after the generation, and they were $P(B|ABA)$ = $0.275\%$ and $P(C|ABA)$ = $0.725\%$.

Next, a target generator was built using the human-generated data for the purpose of producing training and testing patterns used for evaluating the PV-RNN. An MTRNN was used as the target generator by training it with using the human-generated data as the teaching target sequences.
After the training, the closed-loop operation of the MTRNN (feeding next step inputs with current step prediction outputs) generated sample sequence patterns while adding Gaussian noise with zero mean and with constant $\bm{\sigma}$ of 0.05 to the internal state of each context unit at each timestep. This makes the outputs of the network stochastic while preserving the probabilistic structure, not necessarily exactly the same as the one in the training patterns prepared. More details and implementations of this target generator MTRNN can be seen in \textcite{ahmadi2017bridging}. Due to the noise, inserted into the internal dynamics of the MTRNN, the output patterns were noisier and fluctuated more than the human-generated patterns, and those patterns could have different numbers of cycles than $2$. Finally, three groups of patterns were sampled from the MTRNN-generated output patterns, one consisting of $16$ sequence patterns, each with a $400$ step length for the training of the PV-RNN, another comprising $1$ sequence patterns with a $6400$ step length for the first test phase of the PV-RNN, and the last one consisting of $32$ sequence patterns, each with a $400$ step length for the second test phase of the PV-RNN. The main reason that the target generator was used instead of using human-generated trajectory data was because significantly larger number of target data was used (up to 128 sequences) while designing the model, more than could be reasonably created using human generation. The target generator, MTRNN, can effortlessly generate as many instances of patterns as one needs.

\subsection{Experiment 2 with Two Context Layers}
\label{appendix:exp2layers}

Two PV-RNN models were trained with $\wprime$ set to $0.25$. The first model had two context layers consisting of $80$ $\bm{d}$ units and $8$ $\bm{Z}$ units for the fast context (FC) layer, and $40$ $\bm{d}$ units and $4$ $\bm{Z}$ units for the slow context (SC) layer. The time constants of FC and SC units were set to 2 and 4, respectively. The second model had two context layers consisting of $90$ $\bm{d}$ units and $9$ $\bm{Z}$ units for the fast context (FC) layer, and $50$ $\bm{d}$ units and $5$ $\bm{Z}$ units for the slow context (SC) layer. The time constants of FC and SC units were set as in the first model. Training ran for $300,000$ epochs in each case. The first model is equivalent to removing the slow context layer from the PV-RNN model of Section \ref{Exp2} with $\wprime$ set to $0.25$. Moreover, the summations of $\bm{d}$ units and $\bm{Z}$ units in the second model and the PV-RNN model of Section \ref{Exp2} with $\wprime$ set to $0.25$ are equal although the second model has $3640$ more learnable weights. 

The prediction performance using error regression was evaluated for both models as in Section \ref{Exp2}. Tables \ref{Apptab2} and \ref{Apptab3} show the error regression results of the $1$-step to $5$-steps ahead predictions and the $1$-primitive to $3$-primitives predictions, respectively. Both networks show similar results to the network with three context layers (Table \ref{tab4}) for predicting each timestep. However, it can be seen by comparing Table \ref{tab5} with Table \ref{Apptab3} that the network with three context layers outperforms the networks with two context layers for predicting correct patterns.   

\begin{table}[ht]
    \caption{ MSE between the unseen test targets and the $1$-step to $5$-steps ahead generated predictions for models with two context layers.}
    \centering
    \begin{adjustbox}{max width=\textwidth}
    \begin{tabular}{c| c c c c c}
    \cline{1-3}
    \toprule
           \multicolumn{1}{}  & &\multicolumn{5}{c}{Number of Prediction Steps} \\

           \multicolumn{1}{}  &  & 1-step pred.   & 2-steps pred.  & 3-steps pred. & 4-steps pred. & 5-steps pred.\\

    \hline
            First Model& 0.00439  & 0.00912 & 0.014 & 0.0183 & 0.02236\\
    \hline
            Second Model& 0.0041  & 0.00865 & 0.013 & 0.0178 & 0.0223\\
    \bottomrule
    \end{tabular}
    \end{adjustbox}
    \label{Apptab2}
\end{table}

\begin{table}[ht]
    \caption{ Accuracy for $1$, $2$, and $3$ primitive prediction for models with two context layers. }
    \centering
    \begin{adjustbox}{max width=\textwidth}
    \begin{tabular}{c| c c c}
    \cline{1-3}
    \toprule
           \multicolumn{1}{}  & &\multicolumn{3}{c}{Number of Prediction Primitives} \\

           \multicolumn{1}{}  &  & 1-prim. pred. ($\%$)   & 2-prim. pred. ($\%$) & 3-prim. pred. ($\%$)\\

    \hline
            First Model& 90.62  & 75 & 46.87\\
    \hline
            Second Model& 100  & 81.25 & 50\\
    \bottomrule
    \end{tabular}
    \end{adjustbox}
    \label{Apptab3}
\end{table}

\subsection{Experiment 2 with an Alternate Inference Model}
\label{appendix:exp2DiffInfModel}

We examined how providing the past $\bm{d}_{t-1}$ to the inference model could be beneficial by deleting $\bm{d}_{t-1}$ information. The new approximate posterior is obtained as:

\alignedequation{eq:new_posterior_distrib}{
    q_{\phi}(\bm{Z}_{t}~|~\bm{e}_{t:T}) = \mathcal{N}(\bm{Z}_{t}; \bm{\mu}_t^{(q)}, \bm{\sigma}_t^{(q)}) ~~~ \mathrm{where} ~~~ [\bm{\mu}_t^{(q)}, \log{\bm{\sigma}_t^{(q)}}]= f^{(q)}_{\phi}(\bm{A}_{t}^{\widebar{\bm{X}}})
}

A PV-RNN model was trained with $\wprime$ set to $0.25$. Other network's parameters were exactly the same as the PV-RNN models in Section \ref{Exp2}. Table \ref{Apptab4} shows the error regression results of the $1$-step to $5$-steps ahead predictions. By comparing these results with the error regression results of the PV-RNN with $\wprime$ set to $0.25$ shown in Table \ref{tab4}, it can be seen that the model presented here significantly underperforms (it is almost twice as bad) the model with $\bm{d}_{t-1}$ information. 

\begin{table}[ht]
    \caption{ MSE between the unseen test targets and the $1$-step to $5$-steps ahead generated predictions for an inference model without $\bm{d}_{t-1}$ information.}
    \centering
    \begin{adjustbox}{max width=\textwidth}
    \begin{tabular}{c| c c c c c}
    \cline{1-3}
    \toprule
           \multicolumn{1}{}  & &\multicolumn{5}{c}{Number of Prediction Steps} \\

           \multicolumn{1}{}  &  & 1-step pred.   & 2-steps pred.  & 3-steps pred. & 4-steps pred. & 5-steps pred.\\

    \hline
            Alternate Inference Model& 0.0105  & 0.02116 & 0.02977 & 0.03414 & 0.03911\\
    \bottomrule
    \end{tabular}
    \end{adjustbox}
    \label{Apptab4}
\end{table}

\clearpage

\printbibliography
\end{document}